\documentclass{article} % For LaTeX2e
\usepackage{iclr2026_conference,times}

\usepackage{amsmath,amsfonts,bm}

\def\eqref#1{equation~\ref{#1}}
\def\1{\bm{1}}

\DeclareMathAlphabet{\mathsfit}{\encodingdefault}{\sfdefault}{m}{sl}
\SetMathAlphabet{\mathsfit}{bold}{\encodingdefault}{\sfdefault}{bx}{n}

\usepackage{hyperref}
\usepackage{url}
\usepackage{graphicx}
\usepackage{hyperref}
\usepackage{amsmath}
\usepackage{amsfonts} 
\usepackage{float}  
\usepackage{multirow}
\usepackage{lipsum} 
\usepackage{enumitem}
\usepackage{algorithm}
\usepackage{booktabs}
\usepackage{xcolor}
\usepackage{algorithmic}
\usepackage{amssymb}
\usepackage{wrapfig}
\usepackage{booktabs}
\usepackage{multirow}
\usepackage{caption}
\usepackage{array}
\usepackage{setspace}
\usepackage{marvosym}

\definecolor{mygray}{gray}{.9}
\usepackage[table]{xcolor}

\title{Self-signals Driven Multi-LLM Debate for Efficient and Accurate Reasoning}

\author{
    Xuhang Chen\textsuperscript{$1$} \quad
    Zhifan Song\textsuperscript{$2$} \quad
    Deyi Ji\textsuperscript{$3$}  \quad
    Shuo Gao\textsuperscript{$4$\Letter} \quad
    Lanyun Zhu\textsuperscript{$5$\Letter} \\
    $^1$ University of Cambridge \quad
    $^2$ Sorbonne Université  \quad \\
    $^3$ University of Science and Technology of China  \\
    $^4$ Beihang University \quad
    $^5$ Nanyang Technological University \\
}

\iclrfinalcopy % Uncomment for camera-ready version, but NOT for submission.
\begin{document}
\maketitle
\begingroup
\renewcommand\thefootnote{} % 去掉脚注编号
\footnote{\textsuperscript{\Letter}\ Corresponding authors.}
\addtocounter{footnote}{0} % 不影响后续脚注编号
\endgroup

\begin{abstract}
Large Language Models (LLMs) have exhibited impressive capabilities across diverse application domains. Recent work has explored Multi-LLM Agent Debate (MAD) as a way to enhance performance by enabling multiple LLMs to discuss and refine responses iteratively. Nevertheless, existing MAD methods predominantly focus on utilizing external structures, such as debate graphs, using LLM-as-a-Judge, while neglecting the application of self signals, such as token logits and attention, that arise during generation. This omission leads to redundant computation and potential performance degradation. In this paper, we shift the focus to the self signals of multi-LLM debate and introduce a Self-Signals Driven Multi-LLM Debate (SID), which leverages two types of self-signals: model-level confidence and token-level semantic focus, to adaptively guide the debate process. Our approach enables high-confidence agents to exit early at the model level and compress the redundant debate contents based on the attention mechanism. We evaluate our method on various LLMs and Multimodal LLMs across multiple challenging benchmarks. Experimental results demonstrate that our method not only outperforms existing MAD techniques in accuracy but also reduces token consumption, highlighting the effectiveness of utilizing self signals in enhancing both the performance and efficiency of multi-agent debate systems. Our code will be available at~\href{https://github.com/xuhang2019/SID}{\texttt{https://github.com/xuhang2019/SID}}.

\end{abstract}

%\vspace{-0.5\baselineskip}
\section{Introduction}
%\vspace{-0.5\baselineskip}

Large Language Models (LLMs) have demonstrated remarkable capabilities ~\cite{brown2020language, kojima2023largea} across a wide range of domains, including science, technology, engineering, mathematics (STEM) questions ~\cite{hendrycks2021measuring},~\cite{wang2024mmlupro}, and complex reasoning tasks~\cite{rein2023gpqa}. The emergence of Multimodal LLMs (MLLMs) further extends the potential to the visual input domain~\cite{lu2022learn, li2023blip2, liu2024improved}. However, current models still suffer from inherent limitations such as inaccuracies and hallucinations.
%(e.g., factual inaccuracies). 

Multi agent debate (MAD) offers an orthogonal approach to enhancing model performance, in which multiple agents iteratively discuss and refine their answers accordingly~\cite{du2024improving, liu2024breaking, sun2025cortexdebate}. However, a challenge arises from the prevalence of redundant content and repeated consensus points during debate, which not only waste computational resources but also introduce informational noise, potentially impairing the agents' final judgments~\cite{du2024improving, li2024improving}. Moreover, this iterative discussion paradigm incurs substantial token overhead, which becomes increasingly incongruent with the growing capabilities of modern foundation models~\cite{openai2025gptoss120b}. This inherent contradiction between performance gains and token consumption cost presents a central dilemma in contemporary MAD research.

To alleviate this problem, several optimization strategies have been proposed. Broadly, these methods typically fall into two categories: (i) structural optimization, such as adopting various prompting skills~\cite{liu2024breaking}, reducing communication via sparse debate graphs or clustering agents into local debate groups~\cite{liu2024groupdebate}; and (ii) history management, including summarization of prior discussions or introducing agent self-generated confident score~\cite{sun2025cortexdebate}. Whereas these approaches improve the efficiency of information flow in \emph{external} ways (\textit{i.e.,} restructuring agent communication or using LLM-as-a-judge to interpret history), they often suffer from secondary errors such as hallucinations in judges or summaries as evident in~\cite{xiong2023can,zhang2024selfcontrast, tian2025overconfidence}). This limitation motivates us to think: \emph{can we avoid relying on error-prone external mechanisms, and instead leverage more reliable self signals from each agent's generative process to prevent unnecessary and potentially wasteful debate?}

Motivated by the above, in this work, we propose a novel framework that leverages self signals available during LLM inference to improve debate efficiency and performance. In this framework, two types of signals: \textit{model-level confidence} and \textit{token-level semantic focus}, are extracted and used to provide complementary guidance for distinguishing essential information from redundancy, thereby enhancing overall debate quality and efficiency. The model-level confidence, estimated from the probability distribution over the initially generated answer, quantifies how certain the model is about its response. We leverage this signal to design an \textbf{\textit{early-exit mechanism}} that avoids invoking debate when the model is already sufficiently confident, thereby reducing potential noise and redundancy. The token-level semantic focus, derived from attention patterns conditioned on disagreement-oriented prompts, identifies spans in the debate content that the model considers semantically relevant to the disagreement among different agents. We extract and reconstruct these high-attention spans to form a more compact context, thereby introducing a novel \textit{\textbf{compression mechanism}} that preserves critical points of contention while significantly reducing token overhead.

By integrating these two novel mechanisms, each leveraging a different level of self signal, we propose a unified Self \textbf{Si}gnal Driven \textbf{D}ebate framework (SID) to enhance LLM performance. This framework enables early exit for confident agents and extracts focused context for the remaining ones, dynamically adapting the debate process based on the model’s own epistemic signals. We evaluate our method across multiple LLMs and MLLMs on diverse benchmarks, including MMLUpro, Math, GPQA, ScienceQA, and MMstar. SID consistently outperforms existing MAD approaches in most scenarios, while also achieving up to a 40\% reduction in token consumption. These results demonstrate the strong effectiveness of our approach and highlight the significant potential of leveraging internal belief signals in multi-agent systems to jointly optimize performance and efficiency. Our key contributions can be summarized as follows:

\begin{itemize}
    \item We propose a novel idea that leverages self signals from the LLM generation process to enhance agent debate, breaking away from the typical over-reliance on error-prone external mechanisms.
    \item We extract two types of LLM self signals: model-level confidence and token-level semantic focus, and leverage them to design a novel early-exit and a compression mechanism, respectively, effectively reducing redundancy and enhancing debate performance.
    \item Integrating the two proposed mechanisms, we construct an effective and efficient debate framework, SID. Experiments across multiple benchmarks, on both LLMs and MLLMs, demonstrate the significant advantages of SID over existing methods.
\end{itemize}

\section{Related Work}
\label{gen_inst}

\textbf{Reasoning Augmentation} To enhance the reasoning capabilities of LLMs, researchers have explored various techniques. Early work primarily focused on guiding the model through step-by-step reasoning through Chain-of-Thought (CoT) prompts~\cite{zhucpcf, wei2023chainofthought, zhu2025llafs++}  or generating multiple reasoning paths (self-refinement) and voting for the optimal solution through self-consistency or using multi-round self-reflection ~\cite{zhang2024selfcontrast, yao2023tree}. Additionally, subsequent research has found that the model's self-correction capabilities are limited but important \cite{zhu2025retrvr1reasoningdrivenmllmframework}, leading to stagnation in reasoning quality~\cite{zhang2024selfcontrast}. This has partially motivated the rise of multi-agent collaborative paradigms, particularly multi-agent debate (MAD)~\cite{du2024improving}, introducing external perspectives and dynamic feedback among agents to overcome the limits of self-reflection. Our work differs from these studies in that they focus primarily on improving reasoning ability through context prompts, whereas we propose to use self-signals from a model to optimize the context prompt at the token level, thus improving the effectiveness of performance and token ratio.

\textbf{Uncertainty Analysis} 
Uncertainty in LLMs is typically categorized into aleatoric (data-related) and epistemic (model-related) uncertainty\cite{kiureghian2009aleatory, gawlikowski2023survey, zhu2025popen, hu2023uncertainty, ye2025benchmarking}. Given the structured nature of current tasks (e.g., QA, math, science), recent works have focused on quantifying epistemic uncertainty. Mainstream approaches include: (i) probability-based metrics, such as token-level entropy or negative log-likelihood\cite{tu2025ranked}; (ii) ensemble-based methods, e.g., Monte Carlo sampling \cite{metropolis1953equation, hastings1970monte} and Bayesian methods~\cite{kwon2020uncertainty}; and (iii) verbalization-based techniques that prompt the model to self-report confidence~\cite{tian2023just}. Among these, probability-based methods are especially attractive due to their seamless integration with the generation process, without requiring multiple generations, which incur significant token overhead. Recent work by ~\cite{xiong2023can, kirchhof2025position} further explores agent-level uncertainty in interactive settings, emphasizing the role of uncertainty as a confident signal for learning and output control. Our method aligns with the uncertainty in the interactive setting, leveraging self-signals as dynamic indicators of agent-level uncertainty to control agent participation during debates.

\textbf{Multi LLM Debate Systems} Previous multi-LLM debate employs a role-playing setup~\cite{liang2024encouraging}, which has been demonstrated strengths in collaborative tasks. Subsequent research has shown that it is less suited for certain types of problem-solving scenarios. Multi Agent Debate (MAD)~\cite{du2024improving} introduces external perspectives to enrich the system's reasoning capabilities. DMAD~\cite{liu2024breaking} proposes specialized prompt strategies to diversify agent behavior. S2-MAD~\cite{zeng2025s^2mad} introduces a selective sparsity mechanism, allowing agents to selectively participate based on internal cues. CortexDebate~\cite{sun2025cortexdebate} constructs a dynamic sparse debate graph by letting agents serve as self-judges and output confidence scores. These works focus on improving performance via external states (\emph{e.g.,} optimizing structures, using LLM-as-a-Judge). Compared to previous approaches, our method can be orthogonal and complementary, which provides a new angle by integrating self-signals into the debate process instead of optimizing communication structures.

\section{Preliminaries}
We first introduce the naive multi agent debate paradigm in this section. Let $\mathcal{V}$ denote the vocabulary and $\mathrm{Tok}$ the tokenizer. Given a query (e.g., natural language, image, and text) $Q$, $\mathbf{x}=\mathrm{Tok}(Q)$ is the tokenized prompt. A causal LLM $M$ produces a response sequence $\mathbf{y}=(y_1,\dots,y_m)$ with per-step logits $\boldsymbol{\ell}_t\in\mathbb{R}^{|\mathcal{V}|}$ and probabilities $\boldsymbol{\pi}_t=\mathrm{softmax}(\boldsymbol{\ell}_t)$. 
A debate involves $n$ agents $\mathcal{A}=\{1,\dots,n\}$ over rounds $t=0,1,\dots,T$. Let $\mathbf{y}_t^{(j)}$ be agent $j$'s response at round $t$; round $1$ is the initial answering round without debate context. The per-round input to agent $j$ at round $t{+}1$ concatenates the query, its own last response, and other agents' last responses:

\begin{equation}
\label{eq_pre}
\mathbf{x}_{t+1}^{(j)}=\mathrm{Tok}\!\Big(Q \;\Vert\; \mathbf{y}_t^{(j)} \;\Vert\; (Concat_{k\neq j}\mathbf{y}_t^{(k)}) \Big). 
\end{equation}

Here, both $\Vert$ and $Concat$ represent concatenation between prompt groups.

\section{Method}
\label{headings}
\begin{figure}[t]
%\centering
\begin{minipage}{0.55\linewidth}
\vspace{-1.7\baselineskip}
\begin{algorithm}[H]
%\small
\scriptsize
\setstretch{1}
\caption{Self-Signal Driven Debate (SID)}
\label{alg_main}
\begin{algorithmic}[1]
\REQUIRE Query $Q$; LLM agents $\{M_j\}_{j=1}^m$; rounds $N$; confidence threshold $\theta$; top-$p$ ratio $\rho$; 
% \ENSURE Final answer $y^*$
\STATE $Y_1 \leftarrow M_{1}(Q)$ \hfill 
\STATE $u \leftarrow \phi_U(U(Y_1))$ \hfill $\triangleright$ Model-level Confidence(Sec.~\ref{subsection:3.2})
\IF{$u \le \theta$} \RETURN $Y_1$ \ENDIF \hfill $\triangleright$ early exit
\STATE $Y_1 \leftarrow \{y_1\}$;\quad $y^{(j)}_{1} \leftarrow M_j(Q)$ for $j{=}2..m$
\FOR{$t = 2$ to $N$}
  \FOR{$j = 1$ to $m$}
    \STATE $X^{(j)}_{t} \leftarrow \Big(Q \,\Vert\, y^{(j)}_{t-1} \,\Vert\, \texttt{[PROMPT]} \,\Vert\, \text{Concat}_{k \ne j}\big(Y_{t-1}^k\big)\Big)$
    \STATE $A^{(j)}_{t} \leftarrow \textsc{ForwardAttention}\!\big(M_j, X^{(j)}_{t}\big)$ \hfill $\triangleright$ forward only
    \STATE $\widehat{\mathcal{C}}^{(j)}_{t} \leftarrow \mathrm{TopP}\!\big(A^{(j)}_{t}, \rho\big)$
    \STATE $\mathcal{S}^{(j)}_{t} \leftarrow \mathrm{SemanticPreserve}\!\big(\widehat{\mathcal{C}}^{(j)}_{t}\big)$ \hfill $\triangleright$ Sec.~\ref{subsection:3.3}
    \STATE $y^{(j)}_{t} \leftarrow M_j\!\Big(Q \,\Vert\, y^{(j)}_{t-1} \,\Vert\, \mathcal{S}^{(j)}_{t}\Big)$ \hfill $\triangleright$ generate with compressed context
  \ENDFOR
  \STATE $Y_t \leftarrow \{y^{(j)}_{t}\}_{j=1}^m$
\ENDFOR
% \STATE $y^* \leftarrow \phi\!\big(Y_N\big)$ \hfill $\triangleright$ Final Solution
\RETURN $Y_N$
\end{algorithmic}
\end{algorithm}
\end{minipage}
\hfill
\begin{minipage}{0.43\linewidth}
%\vspace{1\baselineskip}
\includegraphics[width=1\linewidth]{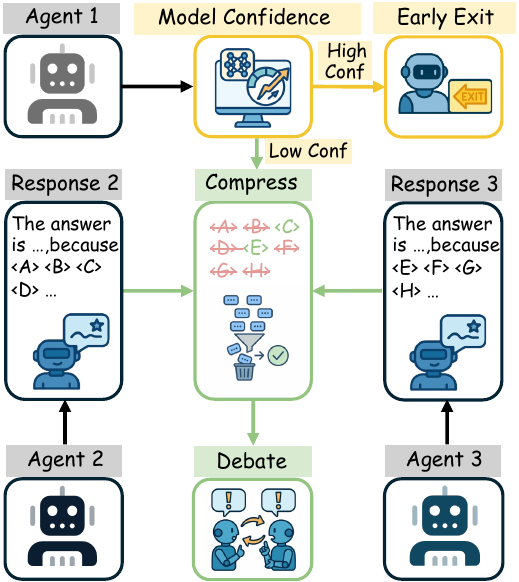}
\vspace{-1\baselineskip}
\caption{Overall framework of SID
}
\label{main_fig}
\end{minipage}
\end{figure}

The above naive framework suffers from several issues, such as excessive redundancy and low efficiency. To address these challenges, as shown in Algorithm~\ref{alg_main} and Figure~\ref{main_fig}, we propose Self Signal Driven Debate (SID), a novel framework that leverages internal confidence signals readily available during inference to adaptively guide the multi-LLM debate process. Specifically, SID utilizes two types of self signals from the LLM: \textit{model-level confidence} and \textit{token-level semantic focus} (see examples in Figure~\ref{fig_case_main_paper} and the Appendix~\ref{app:exa}). \textit{Model-level confidence}, derived from the token-wise output probability distribution (logits), reflects how confident an agent is in its initial answer. We leverage this signal in a newly designed early-exit mechanism to enhance debate efficiency. \textit{Token-level semantic focus}, extracted from the self-attention maps conditioned on disagreement-oriented prompts, captures regions of high variability and knowledge density throughout the debate. This signal is incorporated into a novel compression mechanism to alleviate token redundancy. In the following two sections, we introduce these two components in detail.

\subsection{Early-Exit with Model-Level Confidence}
\label{subsection:3.2}
We first introduce an early-exit mechanism to mitigate redundant debate, motivated by the intuition that cross-LLM discussion is more necessary when a single model lacks confidence in its response. The key to this mechanism lies in extracting an effective confidence score. Intuitively, the more peaked the model’s output distribution over the vocabulary (i.e., lower entropy), the more confident it is in its prediction. Based on this motivation and following conventional methods \cite{tu2025ranked}, we adopt two token-wise uncertainty metrics: entropy $\mathrm{Ent}(\boldsymbol{\pi}_t)=-\sum_{v\in\mathcal{V}}\pi_{t,v}\log \pi_{t,v}$ and negative log-likelihood $\mathrm{NLL}_t=-\log \pi_{t,y_t}$, to estimate the confidence for a generated answer $\mathbf{y}$. To aggregate token-level metrics into a sequence-level confidence score, we explore four aggregation strategies: (1) averaging over tokens (average), (2) taking the maximum value (max), (3) using the first token’s value (first), and (4) using the penultimate token’s value (penultimate). This yields eight confidence measures (four for entropy and four for NLL). Each variant captures different facets of uncertainty: e.g., max focuses on worst-case ambiguity, while penultimate emphasizes late-stage uncertainty often aligned with final reasoning steps in autoregressive generation. 
We concatenate these eight measures to form a vector $U(\mathbf{y})$, which we empirically find to be statistically significant in distinguishing incorrect answers (see Figure~\ref{fig:1-baseline} for details).

 After obtaining confidence scores, the next challenge lies in leveraging them to effectively guide model-level early exits during multi-LLM debates. To this end, we propose two different drop-in strategies that convert the agent’s confidence vector into a binary decision boundary:

\paragraph{Vocabulary-Adaptive Threshold}
 We first tried a straightforward method by directly setting a fixed threshold on the confidence metrics across different types of models. However, this naive strategy yielded suboptimal performance, likely due to the inherent dependency of entropy and NLL magnitudes on the vocabulary size $|\mathcal{V}|$ of the underlying LLM. For example, under a uniform generation assumption, both entropy and NLL equal $\log |\mathcal{V}|$. Thus, larger vocabularies naturally induce higher values, while using a single threshold across models leads to unfairness and unreliability. Based on this motivation, we propose a vocabulary-adaptive threshold as follows:
\begin{equation}
\theta(V)=\alpha\log |\mathcal{V}|,\quad\text{and decide } \mathrm{Terminate}\;\text{iff}\; \phi_{U}(U(\mathbf{y}))\le \theta(V),    
\end{equation}
where $\alpha>0$ is a hyper-parameter, $\phi_{U}$ is an operator to filter noisy metrics $U(\mathbf{y})$. This strategy ensures fair and consistent confidence evaluation across models with varying vocabulary sizes.
%for example, when using attention sink~\cite{xiao2024efficient, openai2025gptoss120b} to modulate generation, the attention pattern is highly concentrated at certain positions, leading to the first and penultimate metrics unreliable.

\paragraph{Calibrated Confidence}  While the aforementioned method provides a simple and robust solution, it relies on a uniformity assumption over token distributions that may not hold in practice. To capture more nuanced confidence signals, we introduce an alternative method, using a lightweight nonlinear classifier $C:\mathbb{R}^d \rightarrow [0, 1]$ trained over a small held-out set. This model takes the confidence vector as input and outputs a scalar confidence score, calibrated against correctness labels:

\begin{equation}
\mathrm{Terminate}\;\text{iff}\; C(U(\mathbf{y}))\ge \tau_c, \quad \tau_c\in(0,1).
\end{equation}

\paragraph{Confidence-Guided Early Exit}
In practice, we adopt the vocabulary-adaptive threshold for gating due to its sufficiently strong performance and training-free simplicity. Specifically, gating is applied in the first round: if an agent’s confidence reaches a high value, it is terminated early, signalling that the system is already sufficiently confident in its answer (see Appendix Figure~\ref{fig:a-ml1}--\ref{fig:a-ml4} for examples). Conversely, if the model exhibits low confidence, this suggests that the question is sufficiently challenging and unlikely to be resolved without additional reasoning, thereby motivating the initiation of a multi-agent debate with the input described in Eq.\ref{eq_pre}.

% We only use gating in the first round because if the LLM is unsure in the first round, this indicates that the problem complexity is high enough to be difficult to solve using the current LLM capabilities without external knowledge input. 
%\begin{figure}[t]

\vspace{-0.25\baselineskip}
\subsection{Adaptive Compression With Token-Level Semantic Focus}
\vspace{-0.25\baselineskip}
\label{subsection:3.3}
In addition to model-level confidence, we further exploit another self signal from the LLM: token-level semantic focus, to improve debate efficiency. As a debate progresses, the accumulated context from multiple agents often becomes repetitive and redundant. We observe that this increasing redundancy can dilute the signal-to-noise ratio, potentially degrading the effectiveness and efficiency of the debate process. A common approach to mitigate this is to use LLM-as-a-judge to summarize past exchanges. However, this method is limited by the summarization capabilities of the base models, which can be prone to hallucination or information loss~\cite{li2024llmsasjudges}. To address this limitation, we instead leverage attention, an intrinsic mechanism of transformer-based models that naturally reflects the model’s focus and the debate's salient region, as an internal signal to implement a token compression framework. %The full concept is illustrated in Figure~\ref{fig:concept}(b).

\paragraph{Prompt-conditioned Attention Extraction}
Given the query $Q$, the agent $j$'s previous answer $\mathbf{y}_t^{(j)}$, and other agents' responses $\{\mathbf{y}_t^{(k)}:k\neq j\}$, we construct a concatenated input:

\begin{equation}
\mathbf{x}_{t+1}^{(j)} = \mathrm{Tok}\left( Q \;\Vert\; \mathbf{y}_t^{(j)} \;\Vert\; \texttt{[PROMPT]} \;\Vert\;  (Concat_{k\neq j}\mathbf{y}_t^{(k)}) 
\right), 
\end{equation}

where \texttt{[PROMPT]} is a task instruction. Here, we use the prompt: 
\texttt{"Identify the key points where they disagree with your own reasoning. Concentrate on those disagreements and decide which line of reasoning is better."}, motivated by prior work demonstrating the benefits of identifying disagreements among agents~\cite{du2024improving, zeng2025s^2mad}. This prompt directs the model’s attention toward segments of the debate that involve semantic conflict, thereby enhancing its focus on critical reasoning divergences. We then define $\mathcal{Q}$ as the set of token positions within the injected prompt, $\mathcal{C}$ as the positions corresponding to other agents’ responses (\emph{i.e.,} $Concat_{k\neq j}\mathbf{y}_t^{(k)}$), and $A^{(l,h)}\in[0,1]^{L\times L}$ as the attention score at layer $l$ and head $h$. For $c\in\mathcal{C}$, a prompt-conditioned semantic focus score is computed by:
\begin{equation}
s(c) = \max_{l,h} \max_{q\in\mathcal{Q}} A^{(l,h)}_{q, c},
\end{equation}

which represents the maximum attention weight from any prompt token to $c$ across all heads and layers, capturing the extent to which $c$ is considered relevant to the disagreement-focused instruction.

\paragraph{Compression with Semantic Preservation}
While token-level attention scores $s(c)$ enable fine-grained identification of salient contents, directly selecting individual tokens may result in fragmented phrases or broken sentence structures. Such fragments hinder the model’s ability to interpret the compressed input coherently. To address this, we apply a semantic preservation heuristic that extends high-attention tokens to complete sub-sentential units. Concretely, we first select the top-$p$ fraction of context tokens, forming $\widehat{\mathcal{C}}=\mathrm{Top}_p\{(c,s(c))\}_{c\in\mathcal{C}}$. Using the tokenizer's offset map $\Psi:c\mapsto [T_a(c), T_b(c)]$, we merge overlapping spans and then expand to sentence boundaries to preserve semantics information. To ensure semantic completeness, we then expand each segment to align with syntactic boundaries, such as commas, periods, or coordinating conjunctions. We denote this process as the $\mathrm{SemanticPreserve}$ operation (see Appendix~\ref{semantic-perserve}, Figure~\ref{fig:a-semantic} for implementation details), which produces a minimal set of semantically coherent text spans as follows:
%We define such operations as $\mathrm{SemanticPreserve}$ operation (please refer to Appendix~\ref{semantic-perserve} for details), and form a minimal set of semantically coherent text spans as follows:

\begin{equation}
\mathcal{S}=\mathrm{SemanticPreserve} \Big(\mathrm{Merge}\big(\{\Psi(c):c\in\widehat{\mathcal{C}}\}\big)\Big).
\end{equation}

We denote the compressed text for agent $j$ corresponding to tokens $S$ as $\mathrm{Text}(\mathcal{S})$, and the next-round input in the debate process (Eq.\ref{eq_pre}) becomes:
\begin{equation}
\widehat{\mathbf{x}}_{t+1}^{(j)}=\mathrm{Tok}\!\Big(Q \;\Vert\; \mathbf{y}_t^{(j)} \;\Vert\; \mathrm{Text}(\mathcal{S})\Big).
\end{equation}

In practice, replacing full histories by $\mathrm{Text}(\mathcal{S})$ yields substantial token compression while preserving points of disagreement.

\vspace{-0.5\baselineskip}
\subsection{Overall Method}
\vspace{-0.5\baselineskip}
    Based on the aforementioned early-exit method with model-level confidence (Sec. \ref{subsection:3.2}) and adaptive compression mechanism with token-level semantic focus (Sec. \ref{subsection:3.3}), we then present the overall SID framework. As shown in Figure~\ref{main_fig}, after initial generation, each agent assesses its confidence using token-level uncertainty metrics derived from output logits. If the agent is sufficiently confident, it exits the debate early, avoiding unnecessary interaction. For less confident cases, the debate proceeds with a compression mechanism guided by the model's own attention dynamics. A disagreement-oriented prompt steers the attention toward semantically relevant spans in other agents’ responses. These spans are then selected and reconstructed into a concise context for the next round, preserving key points of contention.  
By coupling generation-time uncertainty with attention-driven semantic focus, SID adapts the debate trajectory according to each agent’s internal belief state, achieving both high efficiency and robustness without additional training. Readers could refer to  Algorithm~\ref{alg_main} for a more detailed illustration of the overall implementation.

\vspace{-0.5\baselineskip}
\section{Experiments}
\vspace{-0.5\baselineskip}

\subsection{Experiment Setup}
\label{sec:exp-setup}

\paragraph{Tasks and Benchmarks.} Results on both LLM and MLLM tasks are presented. For LLM tasks, we evaluate our method on MMLUpro~\cite{wang2024mmlupro}, 
% GPQA~\cite{rein2023gpqa}, 
and Math~\cite{hendrycks2021measuring} datasets, as they represent a wide range of problem-solving tasks in different domains. For MLLM tasks, we evaluate on ScienceQA~\cite{lu2022learn} and MMStar~\cite{chen2024are} datasets. In consistent with previous methods, we randomly sample $100$ questions from each dataset for evaluation. For the ScienceQA dataset, we utilize the lecture and hint as additional text information following~\cite{liu2024breaking}. For all other datasets, we adopt a zero-shot prompt setting by default.

\paragraph{Models.} To ensure representative coverage of different foundation models, we evaluate both general-purpose and reasoning-oriented models. For LLM tasks, we test on LLaMA-3.1-Instruct-8B (LLaMA3.1-8B) \cite{grattafiori2024llama} and the recently released GPT-OSS-20B \cite{openai2025gptoss120b}. For MLLM tasks, we evaluate LLaVA-v1.6-Vicuna-13B (Hugging Face version, LLaVA1.6-13B) and the GLM4.1V-Thinking (GLM4.1V) reasoning model~\cite{team2025glm45v}.

\paragraph{Implementation Details}
We follow the setup of prior work~\cite{du2024improving, liu2024breaking} to ensure fair comparison, using $n=3$ agents and $N=2$ debate rounds across all SID, MAD, and DMAD settings. The number of self-consistency samples is set to 3. Additionally, we incorporate step-back prompting~\cite{zheng2024take} and self-contrast~\cite{zhang2024selfcontrast} as reasoning augmentation methods in complement to IO (directly output) and COT methods. For model-level confidence, we set the NLL-max threshold $\alpha$ to 1.0 for reasoning-oriented models, 0.5 for general-purpose models, and 0.25 for MLLMs. To mitigate the impact of attention sinks and special tokens on specific token logits~\cite{xiao2024efficient}, we empirically set $\phi(U)$ as the maximum of NLL and entropy, and exclude certain position metrics when computing model-level confidence. The confidence calibration method is trained on a held-out set of 50 samples with $\tau_c$ as 0.9. More implementation details are presented in Appendix~\ref{app:alg}.

\paragraph{Evaluation Metrics}
For the Math dataset~\cite{hendrycks2021measuring}, we adopt the official exact match metric to evaluate agent responses. For all other question-answering datasets, which consist of multiple-choice questions, we use accuracy as the evaluation metric.

\vspace{-0.25\baselineskip}
\subsection{Main Results}
\vspace{-0.25\baselineskip}
% \subsubsection{Performance of SID}

\paragraph{Overall Performance}Table~\ref{table1} and Table~\ref{table_mllm} respectively present the overall performance across LLMs (including LLaMA3.1-8B and GPT-OSS-20B) and MLLMs (including LLaVA1.6-13B model and GLM4.1V) in different datasets. Our SID consistently achieves the best performance in most scenarios, demonstrating its strong effectiveness. Additionally, we observe that MAD methods outperform reasoning augmentation baselines such as self-consistency, which aligns with findings reported in~\cite{liu2024breaking}. Another notable observation is that both the vocabulary-adaptive threshold (SID-v) and calibrated confidence (SID-c) yield very similar performance when implementing the early-exit mechanism described in Sec.\ref{subsection:3.3}. This suggests that the simple thresholding strategy can already approximate the learned decision boundary well. Given its training-free nature and practical effectiveness, we recommend SID-v as the preferred choice in real-world applications.

\begin{figure}[t]
\centering
\includegraphics[width=1.0\linewidth]{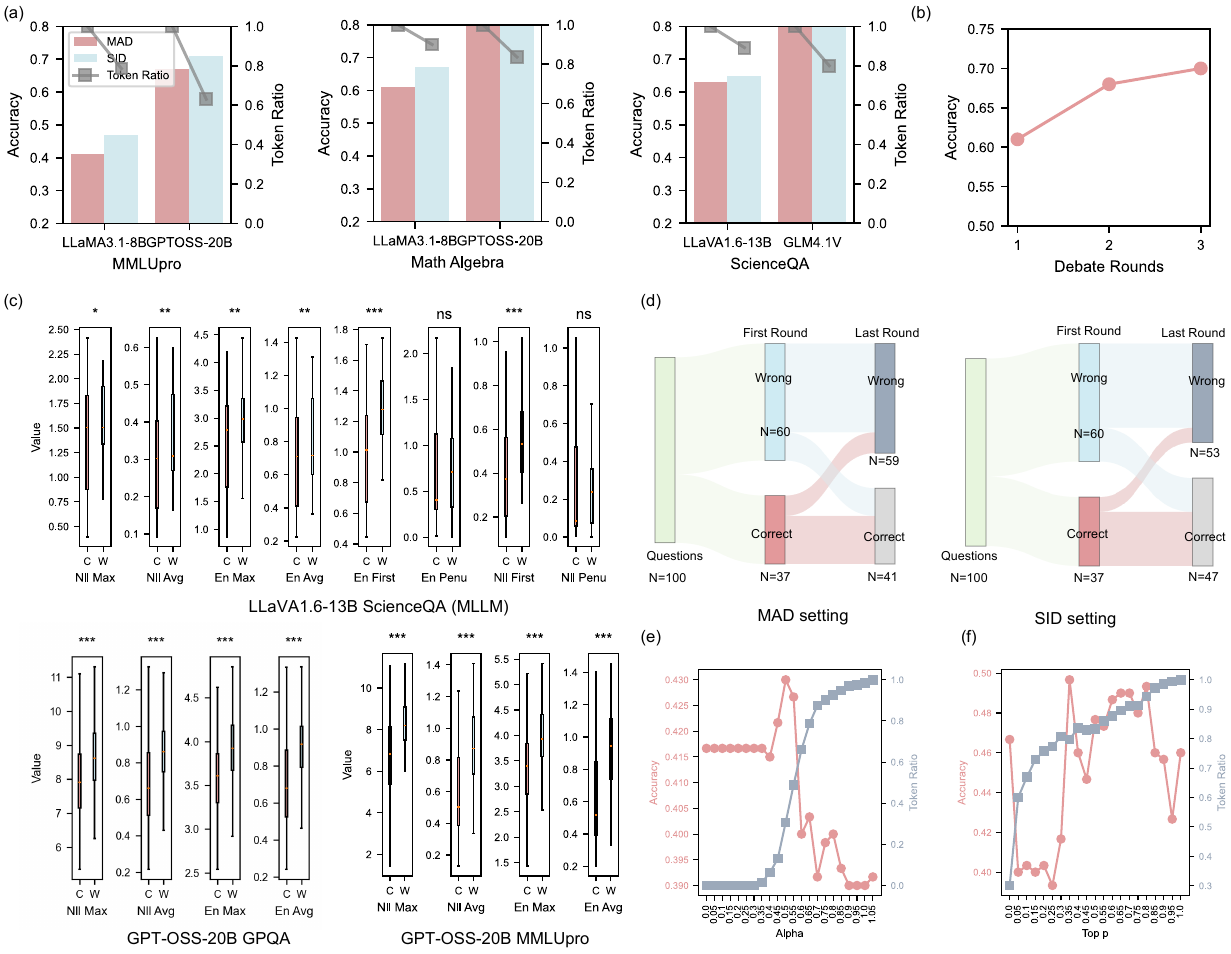}
\caption{(a) Accuracy and token ratio comparison across strategies in MAD vs SID.
(b) Performance with more debate rounds in LLM and MLLM.
(c) Significance tests on model-level confidence signals. C means the correct group, and W means the wrong group. Statistical significance is indicated as follows: $p<0.05$(*), $p<0.01$(**), and $p<0.001$(***):
(d) Answer correction flow in the MAD vs SID setting.
(e) Ablation of $\text{top-}p$ and (f) $\alpha$ on accuracy and token ratio.}
\vspace{-1\baselineskip}
\label{fig:1-baseline}
\end{figure}

\paragraph{Accuracy and Efficiency} 
Figure~\ref{fig:1-baseline}(a) compares the performance and token efficiency of our SID framework against the baseline MAD method, reporting metrics of both the accuracy and the token consumption ratio. The token ratio is computed relative to the MAD setting (i.e., MAD has a token ratio of 1). Results show that SID achieves up to a 40\% reduction in token usage on science and reasoning datasets, while also attaining higher accuracy, demonstrating its significantly better efficiency and effectiveness. Note that on thinking models such as GPT-OSS and GLM4.1V, our method exhibits more significant token reduction, as their reasoning processes are inherently less amenable to token-level compression (see Figure~\ref{fig:a-tl1},\ref{fig:a-tl2} for examples). We also compare the \textit{actual running times} in Figure~\ref{fig:runtime} of the Appendix, where SID demonstrates substantially lower inference time, further underscoring its efficiency advantages. Additionally, Figure~\ref{fig:1-baseline}(b) presents accuracy curves across different debate rounds. SID consistently improves with additional rounds, highlighting its strong scalability under extended deliberation.

\begin{table}[t]
\centering
\caption{Performance comparison across different LLMs for various datasets (Math subsets and MMLUpro). SID-v and SID-c denote our method using the vocabulary-adaptive threshold and calibrated confidence, respectively, to implement the early-exit mechanism. (see Sec.\ref{subsection:3.2} for details)}
%\vspace{-15pt}
\resizebox{1.0\columnwidth}{!}{
\label{table1}
\setlength\tabcolsep{8pt}
\begin{tabular}{llccccccccc}
\toprule
Model & Method & Alg. & C\&P & Geo. & Int.A.& Num & Pre.A &Pre.C. & MMLUpro & Avg  \\
\midrule
% \addlinespace 
% first area (Upper area)
\multirow{9}{*}{LLaMA3.1-8B} 
& COT & 61 & 38 & 34 & 14 & 37 & 54 & 28 & 39 & 38.13  \\
& IO & 65 & 37 & 35 & 15 & \textbf{46} & 59 & 28 & 25 & 38.75 \\
& SBP~\cite{zheng2024take} & 46 & 28 & 21 & 12 & 33 & 46 & 24 & 15 & 28.13\\
& Self-Consistency~\cite{wang2023selfconsistencya} & 58 & 25 & 32 & 12 & 40 & 55 & 25 & 45 & 36.50 \\
& Self-Contrast~\cite{zhang2024selfcontrast} & 54 & 36 & 27 & 11 & 31 & 53 & 27 & 36 & 34.38\\
& MAD~\cite{du2024improving} & 61 & 36 & 36 & 16 & 37 & 60 & 29 & 41 & 39.50\\
& DMAD~\cite{liu2024breaking} & 55 & 36 & 32 & 13 & 36 & 58 & 26 & 39 & 36.88\\
\cmidrule(lr){2-11} 
\rowcolor{mygray}& SID-v & \textbf{67} & \textbf{43} & \textbf{40} & 18 & 41 & 64 & \textbf{31} & \textbf{47} & 43.88\\
\rowcolor{mygray}& SID-c  & \textbf{67} & \textbf{43} & 39 & \textbf{20} & 41 & \textbf{65} & 30 & \textbf{47} & \textbf{44.00}\ \\
\midrule

% second area
\multirow{8}{*}{GPT-OSS-20B} 
& COT & 85 & 81 & 56 & 36 & 70 & 84 & 44 & 61 & 64.63 \\
& IO & 85 & 81 & 60 & 40 & 74 & 87 & 42 & 64  & 66.63\\
& SBP~\cite{zheng2024take} & 65 & 64 & 44 & 37 & 16 & 73 & 11 & 26 & 42.00  \\
& Self-Consistency~\cite{wang2023selfconsistencya} & 75 & 67 & 44 & 31 & 70 & 79 & 23 & 69 & 57.25\\
& Self-Contrast~\cite{zhang2024selfcontrast} & 84 & 75 & 65 & 36 & 67 & 88 & 35 & 65 & 64.38\\
%& MAD~\cite{du2024improving} & \textbf{95} & 91 & \textbf{80} & \textbf{67} & \textbf{91} & \textbf{62} & 67 \\
& DMAD~\cite{liu2024breaking} & 91 & 90 & 73 & 51 & 66 & 89 & 47 & 65 & 71.50\\
\cmidrule(lr){2-11}
\rowcolor{mygray}&SID-v & \textbf{94} & \textbf{92} & 79 & \textbf{65} & \textbf{87} & \textbf{91} & \textbf{62} & \textbf{71} & \textbf{80.13} \\
\rowcolor{mygray}& SID-c & \textbf{94} & \textbf{92} & \textbf{80} & 62 & \textbf{87} & \textbf{91} & 61 & 70 & 79.63\\
\bottomrule
\end{tabular}
}
\end{table}

\paragraph{Statistical Significance Analysis}
The statistical significance of our model-level confidence metric is illustrated in Figure~\ref{fig:1-baseline}(c) and Figure~\ref{fig:c-llama-algebra}--\ref{fig:c-gpt-gpqa}, where results for both the LLM (GPT-OSS-20B) and MLLM (LLaVA1.6-13B) are presented. In the figure, C and W denote correct and incorrect responses, respectively. Across two tasks of varying difficulty: GPQA and MMLUpro, our SID maintains a consistent confidence threshold within the correct group for the same model (e.g., NLL max $\approx 7.5$), highlighting the stability and robustness of our model-level confidence signal.

\begin{table}[t]
\begin{minipage}{0.55\linewidth}
\setlength{\tabcolsep}{1pt}
\centering
\caption{Performance on Sci.QA and MMStar based on MLLMs LLaVA1.6-13B and GLM4.1V.}
\resizebox{\linewidth}{!}{
\renewcommand{\arraystretch}{1.12}
\begin{tabular}{llcc|llcc}
\toprule
% \multicolumn{4}{c|}{\textbf{(a) LlaVA1.6-13B}} & \multicolumn{4}{c}{\textbf{(b) GLM4.1V}} \\
% \cmidrule(lr){1-4} \cmidrule(lr){5-8}
Model & Method & Sci.QA & MMStar & Model & Method & Sci.QA & MMStar \\
\cmidrule(lr){1-8}
\multirow{6}{*}{LLaVA1.6-13B} & CoT & 63 & 11 & \multirow{6}{*}{GLM4.1V} & CoT & 83 & 29\\
& IO & 62 & 9 & & IO & 83 & 32 \\
% & SBP & 67 & 16 & & SBP & 84 & 55 \\
& Self-Consis& 63 & 11 & & Self-Consis & 84 & 29 \\
%& Self-Contrast & 59 & 27 & & Self-Consistency & 90 & 64 \\
% & DDCoT & 64 & 28 & & DDCoT & 89 & 44 \\
% & CCoT & 67 & 15 & & CCoT & 85 & \textbf{58} \\
& MAD & 65 & 12 & & MAD & 90 & 47 \\
%& DMAD & \textbf{65} & 24 & & DMAD & 89 & 60 \\
% & DMAD & 20 & tbd & & DMAD & tbd & tbd \\
\cmidrule(lr){2-4} \cmidrule(lr){6-8}
\rowcolor{mygray}& SID-v & \textbf{65} & \textbf{14} & & SID-v & \textbf{91} & \textbf{54}\\
\rowcolor{mygray}& SID-c & \textbf{65} & \textbf{14} & & SID-c & \textbf{91} & \textbf{54} \\
\bottomrule
\end{tabular}
\label{table_mllm}
}
\end{minipage}
\hfill
\begin{minipage}{0.44\linewidth}
\setlength{\tabcolsep}{2pt}
\centering
\caption{Ablation Study of SID on MMLUpro based on LLaMA3.1-8B.}
\resizebox{\linewidth}{!}{
\renewcommand{\arraystretch}{1.04}
\begin{tabular}{lcc}
\toprule
Method & Accuracy & Token Ratio \\
\midrule
Baseline Single-round CoT & 37.67 & 0.17\\
Baseline MAD & 39.50 & 1.00 \\
Baseline MAD + Compression & 41.67 & 0.73 \\
Baseline MAD + Compression + Early Exit & 46.83 & 0.53\\
\toprule
SID w/o Semantic Preservation & 34.50 & 0.46 \\
SID w/o Early Exit w/ Token-level Summary & 39.50 & 0.68 \\
SID & 46.83 & 0.53\\
\bottomrule
\end{tabular}
\label{table_ablation}
}
\end{minipage}
\end{table}

\vspace{-0.25\baselineskip}
\subsection{Ablation and Analysis}
\vspace{-0.25\baselineskip}

\paragraph{Ablation of Key Components} Using the LLaMA3.1–8B model and the MMLUpro dataset, we conduct a comprehensive ablation study to evaluate the key design components of our framework. As shown in Table~\ref{table_ablation}, the baseline MAD setup yields suboptimal performance. In contrast, incorporating our proposed early-exit mechanism based on model-level confidence (Section~\ref{subsection:3.2}) and the compression mechanism guided by token-level semantic focus (Section~\ref{subsection:3.3}) leads to substantial improvements in both effectiveness and token efficiency. We further evaluate semantic preservation in our compression mechanism that helps enforce the completion of sub-sentential units when extracting semantic focus information. The significant performance degradation observed when this component is excluded highlights its importance and effectiveness. Additionally, replacing token-level compression with a self-summary strategy, a representative form of the LLM-as-a-judge paradigm, is also evaluated for comparison. As shown in Table~\ref{table_ablation}, this approach yields slightly worse results than ours, indicating that the self-signals within LLM’s generation process can help enhance the debate more efficiently and robustly. These results validate the design choices of our framework and highlight the contribution of each component to the overall performance.

\vspace{-0.5\baselineskip}
\paragraph{Ablation of Vocabulary Adaptive Threshold $\alpha$} We further conduct an ablation study on the vocabulary adaptive threshold $\alpha$ and early exit ratio based on the LLaMA3.1-8B. The results are presented in  Figure~\ref{fig:1-baseline} (e). Small $\alpha$ means all questions are unconfident, thus the exit ratio is 0, equivalent to traditional MAD, whereas large $\alpha$ means all questions are confident, thus the system stops at the first round, equivalent to only one LLM model. Our results show that $\alpha = 0.5$ is an optimal value for this LLaMA3.1-8B model. 

\vspace{-0.5\baselineskip}
\paragraph{Ablation of Semantic Preservation Ratio $P$ }In our semantic preservation framework, we select the top‑$p$ fraction of context tokens for further processing. The ablation study results for varying $p$ are shown in Figure~\ref{fig:1-baseline} (f). We observe that selecting the top tokens with $p$ around 0.35 or 0.4 yields the best performance. It is interesting to find that when $p > 0$ but very small, performance can degrade compared to the case where no additional context is included. Conversely, when $p$ is too large, which means retaining a broader range of content, including potentially redundant agreement, the performance also drops. These findings suggest that both incomplete and overly redundant context can negatively impact multi-LLM debate effectiveness.

\begin{figure}[!t]
\centering
\includegraphics[width=1\linewidth]{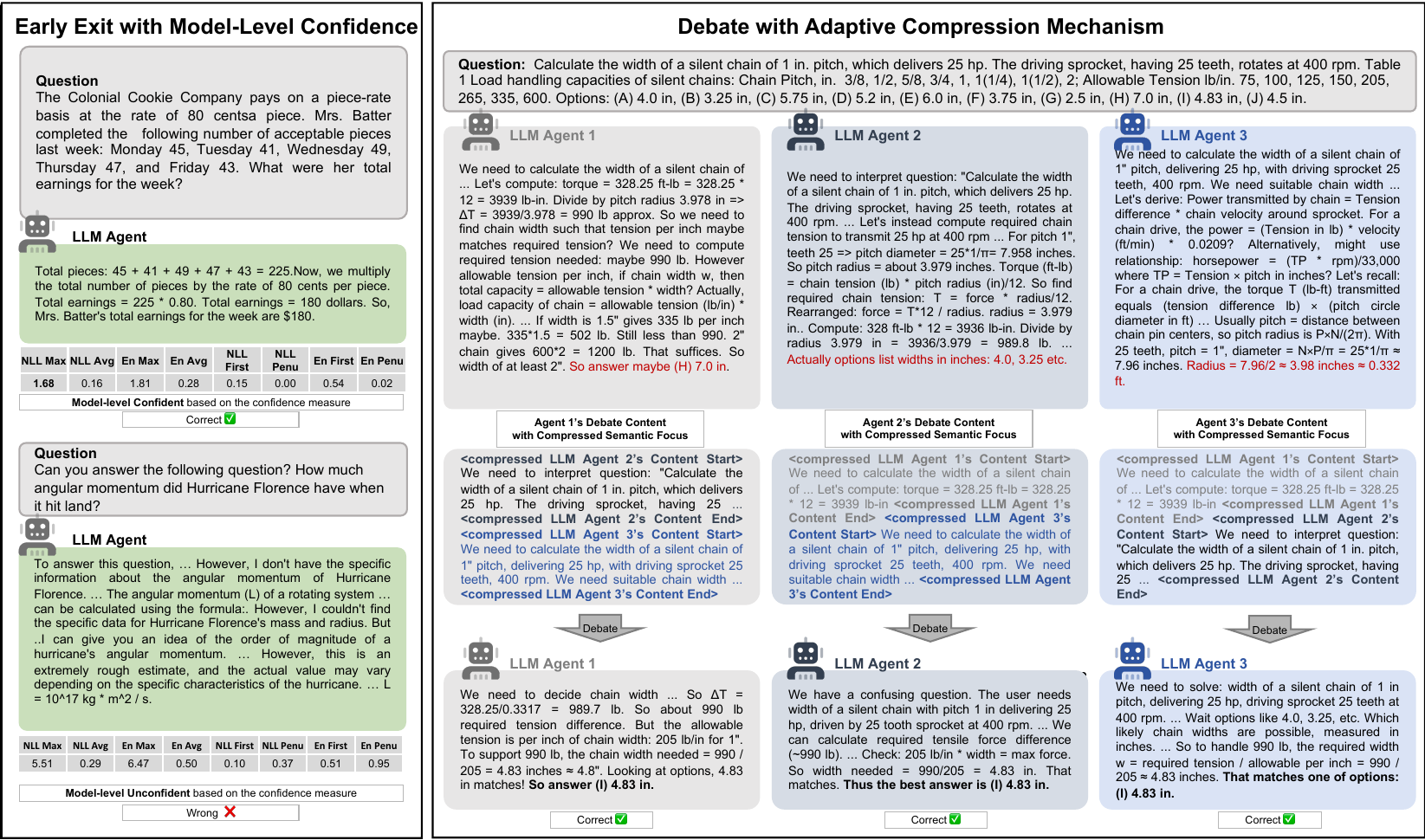}
\caption{Case study of SID's debate process. (Left) On MMLUpro, SID exits early for a simple arithmetic question with high confidence but fails on a complex physics question with low confidence. (Right) Three agents initially err but converge to the correct answer through debate guided by token-level semantic focus from adaptively compressed content.
}
\label{fig_case_main_paper}
\end{figure}

%\vspace{-0.5\baselineskip}
\subsection{Visualization Results}
%\vspace{-0.5\baselineskip}
To illustrate the mechanisms of our framework more intuitively, we present the visualizations of SID’s workflow in Figure~\ref{fig_case_main_paper}. The left branch showcases the early-exit mechanism on a real-world economics question. After generating an answer, the model is assessed as highly confident (e.g., NLL Max = 1.68) by the model-level confidence module and exits early with a correct prediction. In contrast, for a more complex physics question, the model is flagged as low confidence (e.g., NLL Max = 5.51), thus prompting further debate. The right branch illustrates the debate process guided by our adaptive compression mechanism. When facing a challenging physics problem, all three agents initially fail. However, by engaging in a debate using token-level compressed content driven by semantic focus, the agents collaboratively refine their reasoning and successfully converge on the correct answer. More case studies can be found in Appendix~\ref{app:exa}, Figure~\ref{fig:a-tl2}. Furthermore, Figure~\ref{fig:1-baseline}(d) compares the corrections made by the MAD and SID. Our method significantly reduces the number of cases where debates drift from correct to incorrect answers, while increasing the number of beneficial corrections, i.e., debates that shift from wrong to correct outcomes, further demonstrating the high effectiveness of our method. 

%\vspace{-0.5\baselineskip}
\section{Conclusion}
%\vspace{-0.5\baselineskip}
This work introduces SID, a multi-LLM debate framework that leverages self signals from the LLM generation process to improve both performance and efficiency. SID integrates two types of internal signals: model-level confidence, which enables early exit for confident agents, and token-level semantic focus, which compresses debate history by using attention scores to retain key points of disagreement. Experiments across diverse benchmarks with various LLMs and MLLMs demonstrate the high performance and efficiency of SID, underscoring the strong potential of leveraging internal model states as effective signals for guiding collaborative problem-solving. These findings point toward a promising direction for developing new paradigms in multi-agent systems.

%This work introduces SID, a multi-LLM debate framework that leverages self signals from the LLM generation process to improve both performance and efficiency. SID integrates two types of self signals: a model-level confidence to enable early exit for confident agents, and a token-level semantic focus that helps compress debate history using attention scores to retain key points of disagreement. Experiments across diverse benchmarks with various LLMs and MLLMs demonstrate that internal model states can serve as effective signals for guiding collaborative problem-solving. Our findings suggest a promising direction for developing novel paradigms for multi-agent system. 

%integrating internal dynamics into future multi-agent systems.

\clearpage
\section*{Reproducibility statement}
Significant efforts have been made to ensure the reproducibility of our results. The implementation details of our framework are described in the main manuscript (Section~\ref{headings}, Algorithm~\ref{alg_main}, and \ref{sec:exp-setup}), including methods, baselines, benchmarks, model configurations, and evaluation settings. Additional implementation details and the full algorithm are provided in Appendix~\ref{app:alg}. To facilitate faithful replication of our method, we include detailed descriptions of the key prompts and instruction formats in Table~\ref{tab:system-prompts}, Table~\ref{tab:instruction-format}, and Figure~\ref{fig:a-reasoning-prompt}. We believe these materials are sufficient to enable reproducibility of our study.

\bibliography{iclr2026_conference}
\bibliographystyle{iclr2026_conference}

\clearpage
\appendix
\section{Use of Large Language Models (LLMs)}
Large Language Models (LLMs) were used solely for language refinement and proofreading purposes. They were not involved in research ideation and methodology design. All scientific contributions and conceptual developments were carried out entirely by the authors. The LLM did not play a substantive role in shaping the research content and should not be considered a contributor.

\section{Limitation}
\label{app:lim}
Our method relies on internal model signals such as logits and attention maps, which limit direct applicability to public closed-source APIs. However, it remains well-suited for internal deployments of proprietary models, especially in multi-agent systems, and can serve as an intermediate reasoning layer prior to externalized API serving. Notably, many modern systems (e.g., GPT-5) already adopt multi-agent or tool-augmented architectures, making our approach broadly applicable and increasingly relevant.

\section{Algorithm and Implementation}
\label{app:alg}

\subsection{Semantic Preservation}
\label{semantic-perserve}

The semantic preservation module plays a key role in restoring the semantic cohesion from the top-$p$ selected sparse tokens based on a model's self-signals (\emph{i.e.,} attention mechanism). Specifically, the method selects the most relevant textual spans based on attention distribution, but ensures that these selections are semantically coherent when mapped back to natural language. The algorithm below (Algorithm~\ref{alg:semantic-preserve}) shows the main pipeline, and the example (Figure~\ref{fig:a-semantic}) illustrates the comparison between without and with semantic preservation.

\begin{algorithm}[H]
\caption{Semantic-Preserving Compression}
\label{alg:semantic-preserve}
\begin{algorithmic}[1]
\REQUIRE Prompt text $x$ with marked spans \texttt{FOCUS}, \texttt{DISCUSSION}; offset map $\mathcal{O}$; Top-p selected attention score $\mathcal{C}$  
\ENSURE Compressed prompt $x'$
\STATE $\mathcal{U} \leftarrow \textsc{ExtractUnits}(x, \texttt{DISCUSSION})$ \hfill $\triangleright$ Sentence/clause-level segments
\STATE $T \leftarrow \textsc{Tokenizer}(x)$
\STATE $\mathcal{S} \leftarrow \textsc{MapTokensToUnits}(\mathcal{C}, \mathcal{U}, \mathcal{O})$
\STATE $x' \leftarrow \textsc{ReplaceSpan}(x, \texttt{DISCUSSION}, \mathcal{S})$
\RETURN $x'$ 
\end{algorithmic}
\end{algorithm}

We begin by extracting semantically coherent units (e.g., sentences or clauses) from the \texttt{DISCUSSION} span using lightweight parsing heuristics, including punctuation or newline segmentation. This yields a set of candidate text fragments $\mathcal{U}$.

We then use the Top-p selected attention score $\mathcal{C}$ (from Algorithm~\ref{alg_main}) to select the top-$p$ most relevant tokens from T. To preserve semantic interpretability, we map these selected tokens back to their enclosing segments in $\mathcal{U}$ using the token-to-text offset map $\mathcal{O}$. The resulting set of informative fragments $\mathcal{S}$ is used to replace the original \texttt{DISCUSSION} span, yielding a compressed prompt $x'$ that retains critical disagreement signals while discarding redundant or low-relevance content.

In the multi-modal setting (e.g., MLLMs), token offsets may shift due to image-text fusion. We mitigate this by anchoring to stable textual markers in the \texttt{FOCUS} span to adjust $\mathcal{O}$ and maintain alignment. 

This compression module is integrated into the overall SID framework to support efficient and interpretable multi-agent reasoning under token or latency constraints

\subsection{Prompt Template}

In multi-task evaluation settings, especially those involving factual or multiple-choice benchmarks, we observe that models frequently generate semantically correct answers but fail to conform to the expected output format. This discrepancy is particularly pronounced in open-ended LLMs, where prior supervised fine-tuning (SFT) phases may introduce implicit formatting preferences (e.g., \texttt{\textbackslash boxed} in math domains).

To mitigate this, we prepend a task-specific \textit{system prompt} that explicitly enforces the desired answer format. Our full prompting format is:
\begin{equation*}
    \texttt{<system prompt>} + \texttt{<question content>} + \texttt{<output instruction>}
\end{equation*}
This method proves especially helpful for models with weaker instruction-following capabilities (e.g., LLaMA3.1-8B) and significantly reduces post-hoc answer parsing failures.  Another example is the GLM4.1-V thinking model. The default multiple choice response uses a special boxed token, such as  \verb+<|begin_of_box|>B<|end_of_box|>+.
By emphasizing the answer returning with brackets in the system prompt, GLM4.1V thinking yields  \verb+<|begin_of_box|>(B)<|end_of_box|>+.  This allows us to extract the result using brackets in a unified way.Table~\ref{tab:system-prompts} lists the dataset-specific system prompts and the enforced answer formats used in our experiments.

\begin{table}[h]
\centering
\caption{Dataset-specific system prompts and enforced output formats for answer extraction.}
\label{tab:system-prompts}
\begin{tabular}{l|p{7cm}|p{4cm}}
\hline
\textbf{Dataset} & \textbf{System Prompt (Instruction)} & \textbf{Expected Output Format (for answer parsing)} \\
\hline
\texttt{MMLUpro} & You are a trivia expert who knows everything. You are tasked to answer the following question. Give your final answer in the format of \texttt{(X)}, e.g., \texttt{(A)}. & \texttt{(A)}, \texttt{(B)}, etc. \\
\hline
\texttt{Math} & You are a math expert. You are tasked to determine the answer to the following question. Give your final answer in the form of \texttt{\textbackslash boxed\{answer\}} in the last sentence of your response, e.g., \texttt{\textbackslash boxed\{[1, 3]\}}. & \texttt{\textbackslash boxed\{...\}} \\
\hline
\texttt{GPQA} & You are an expert in graduate-level science and mathematics. You will be presented with challenging questions designed to test your reasoning abilities. Your last sentence should be ``The correct answer is \texttt{(insert answer here)}.'' & ``The correct answer is \texttt{(A)}.'' \\
\hline
\texttt{ScienceQA} & You are a trivia expert who knows everything. You are tasked to answer the following question. Give your final answer in the format of \texttt{(X)}, e.g., \texttt{(A)}. & \texttt{(A)}, \texttt{(B)}, etc. \\
\hline
\texttt{MMStar} & You are an expert in multimodal task understanding, and your task is to answer the following questions. Give your final answer in the format of (X), e.g., (A) & \texttt{(A)}, \texttt{(B)}, etc. \\
\hline
\end{tabular}
\end{table}

\begin{table}[H]
\centering
\caption{Dataset-specific output instruction prompts.}
\label{tab:instruction-format}
\begin{tabular}{c|p{9cm}}
\hline
\textbf{Dataset} & \textbf{Output Instruction} \\
\hline
\texttt{MMLUpro} & Give your final answer in the format of '(X)'\\
\hline
\texttt{Math} & Give your final answer in the form of \verb|\\boxed{answer}| at the end of your response, e.g., \verb|\\boxed{[1, 3]}|.\\
\hline
\texttt{GPQA} & Your last sentence should be 'The correct answer is (insert answer here).' e.g., The correct answer is (A).\\
\hline
\texttt{ScienceQA} & Give your final answer in the format of '(X)'. You should only give one answer. For example, the answer is (A).\\
\hline
\texttt{MMStar} & Give your final answer in the format of '(X)'. You should only give one answer. \\
\hline
\end{tabular}
\end{table}

In terms of question content, we strictly follow the previous work~\cite{du2024improving, liu2024breaking} in parsing the question to the chat template.

\begin{figure}[!t]
\centering
\includegraphics[page=1, width=0.95\linewidth]{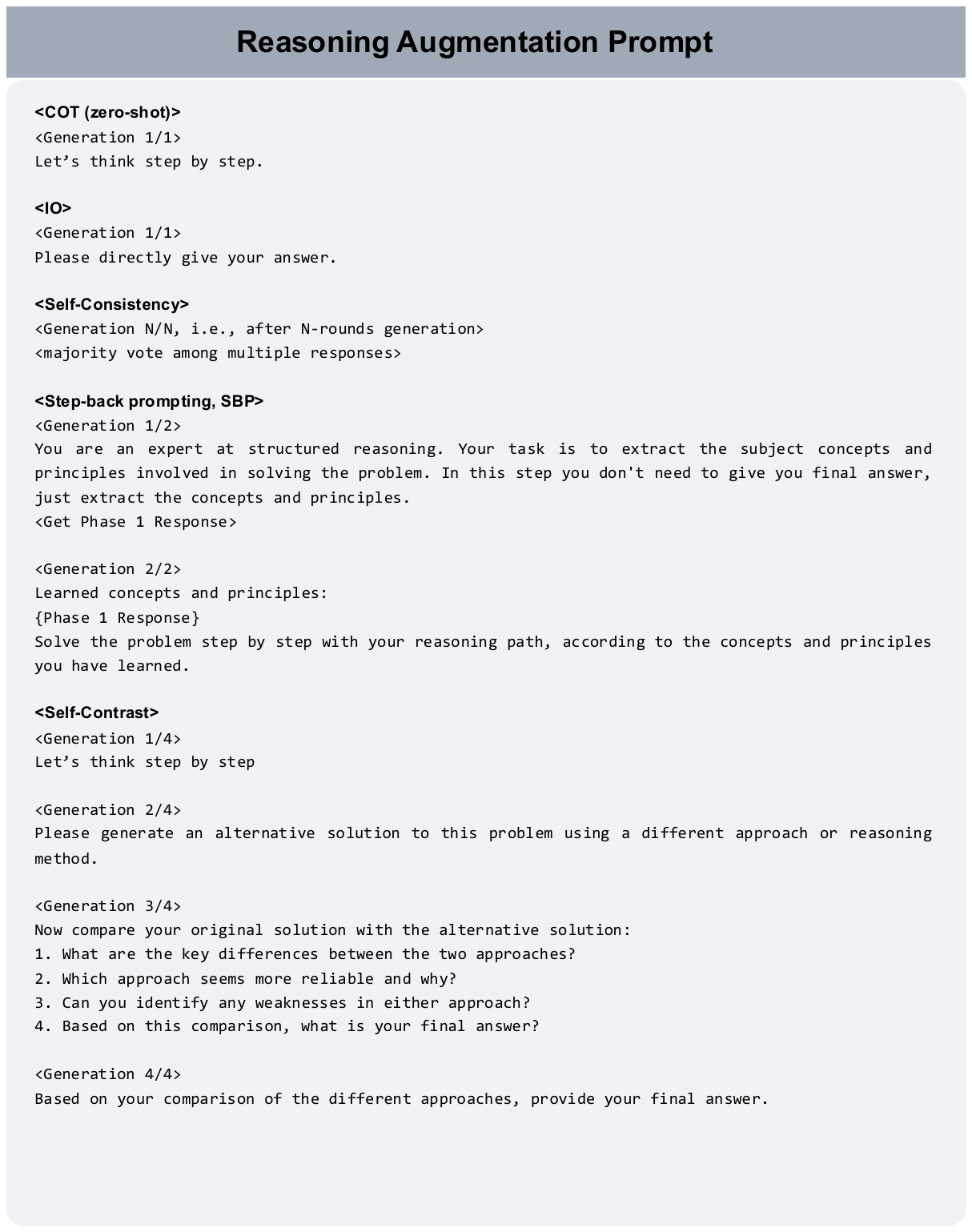}
\caption{Details of reasoning augmentation prompt.}
\label{fig:a-reasoning-prompt}
\end{figure}

Moreover, we list the reasoning augmentation prompt (Figure~\ref{fig:a-reasoning-prompt} used in our experiments. Notably, Output Instructions should still be used after those prompts to enhance the ability to follow instructions.

\clearpage
\section{More Results}
\label{app:exa}

In this section, we list many visualization results to illustrate the effectiveness of our SID methods.

\begin{figure}[!t]
\centering
\includegraphics[width=0.98\linewidth]{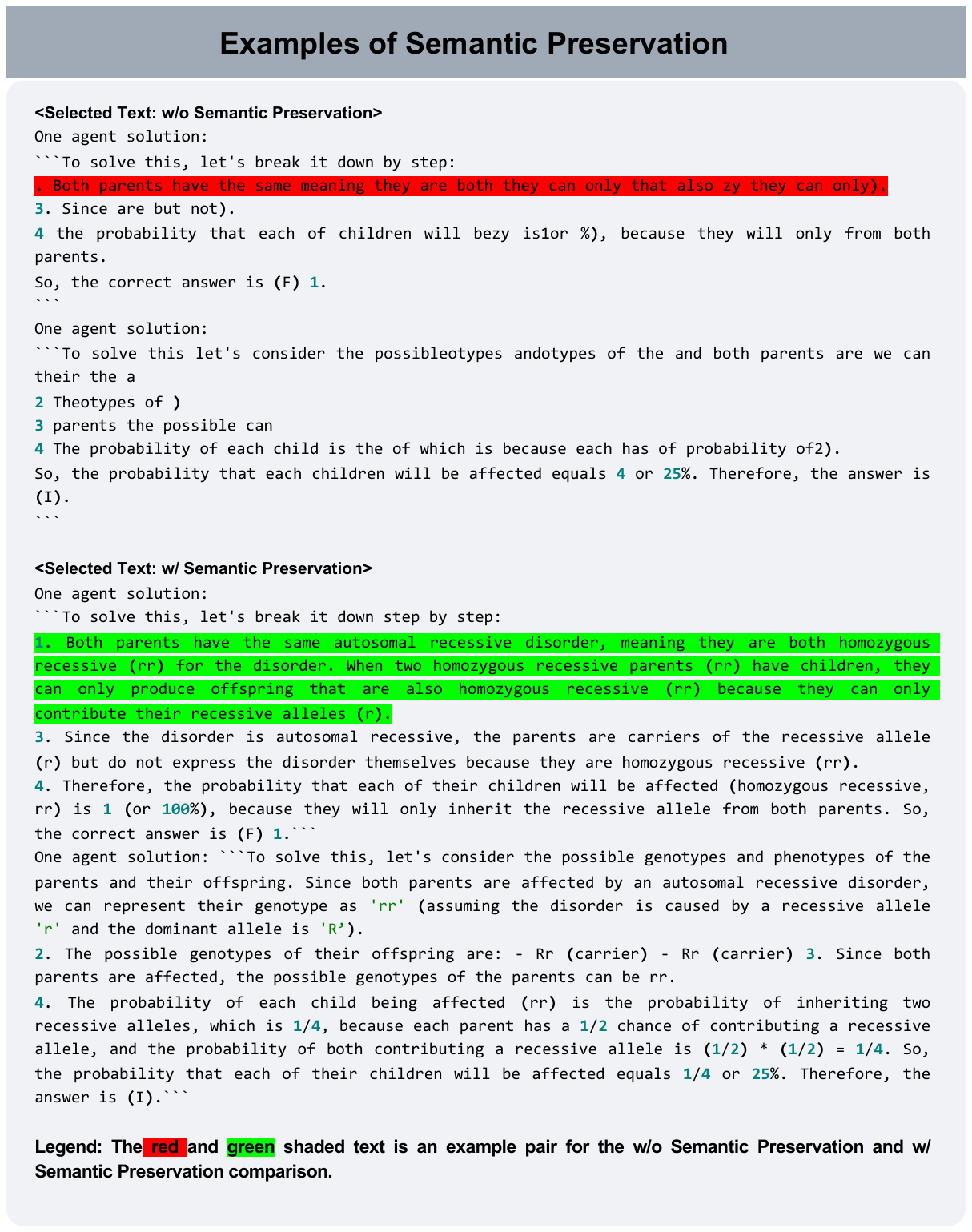}
\caption{Example comparison between w/o semantic preservation (red, brute-force token selection) and w/ semantic preservation (green, semantically coherent expansion)}
\label{fig:a-semantic}
\end{figure}

Figure~\ref{fig:a-semantic} Example comparison between w/o semantic preservation (red, brute-force token selection) and w/ semantic preservation (green, semantically coherent expansion). It can be observed that token-level semantic focus deletes irrelevant points (for example, point 2 in the solution was deleted), and our semantic preservation retains the semantic cohesion from the selected tokens.

A series of model-level confidence examples below can demonstrate the stable early exit threshold in the same model, and the statistical significance between the correct and wrong groups. Moreover, we also provide the correction flow from the first round to the last round.

\begin{figure}[!t]
\centering
\includegraphics[width=0.95\linewidth]{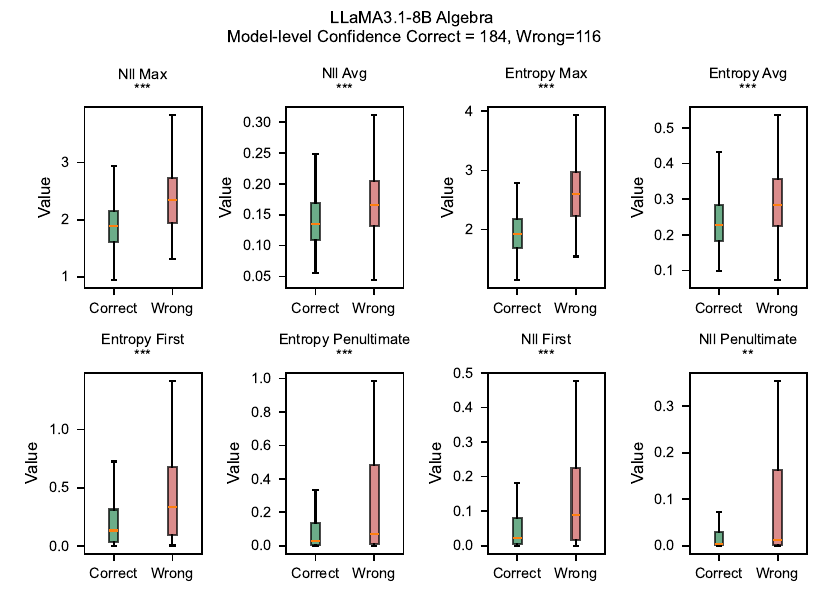}
\vspace{0.1cm} 
\includegraphics[width=0.95\linewidth]{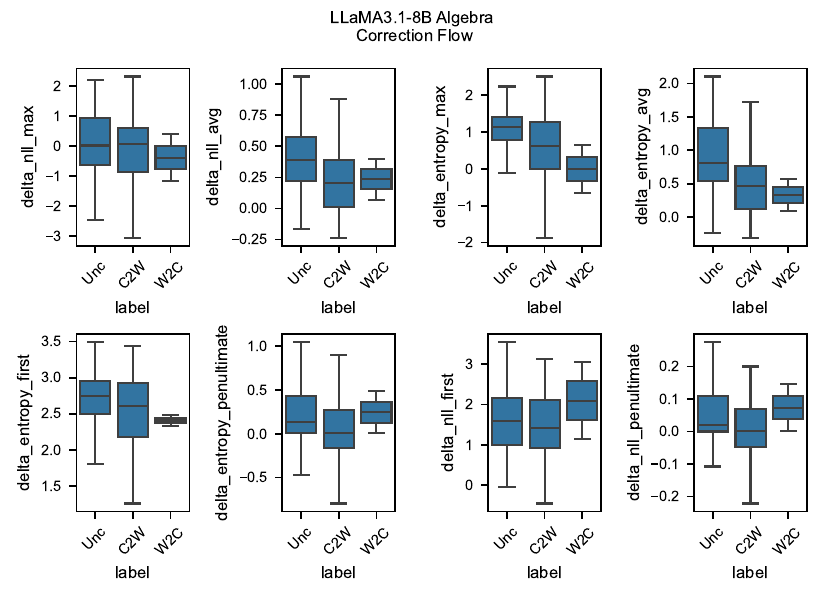}
\caption{Top: Model-level Confidence result on the Math Algebra dataset with LLaMA3.1-8B. Bottom: Correction flow with 8 deltas of confidence metrics. Unc: the answer remains unchanged between the first round and the last round, C2W: correct to wrong, W2C: wrong to correct.}
\label{fig:c-llama-algebra}
\end{figure}

Figure~\ref{fig:c-llama-algebra} Top: Model-level Confidence result on the Math Algebra dataset with LLaMA3.1-8B. Bottom: Correction flow with 8 deltas of confidence metrics. Unc: the answer remains unchanged between the first round and the last round, C2W: correct to wrong, W2C: wrong to correct.

\begin{figure}[!t]
\centering
\includegraphics[width=0.95\linewidth]{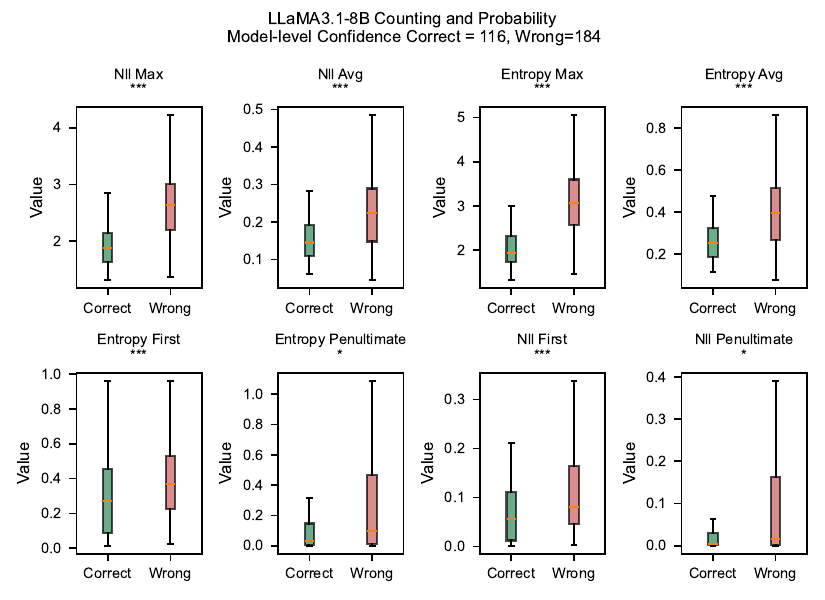}
\vspace{0.1cm} 
\includegraphics[width=0.95\linewidth]{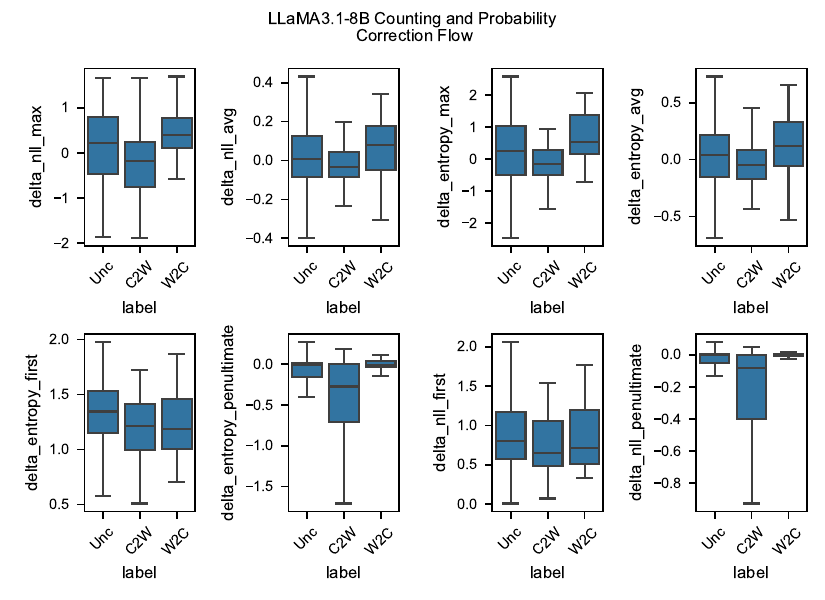}
\caption{Top: Model-level Confidence result on the Math Counting and Probability dataset with LLaMA3.1-8B. Bottom: Correction flow with 8 deltas of confidence metrics. Unc: the answer remains unchanged between the first round and the last round, C2W: correct to wrong, W2C: wrong to correct.}
\label{fig:c-llama-cp}
\end{figure}

Figure~\ref{fig:c-llama-cp} Model-level Confidence result on the Math Counting and Probability dataset with LLaMA3.1-8B. 

\begin{figure}[!t]
\centering
\includegraphics[width=0.95\linewidth]{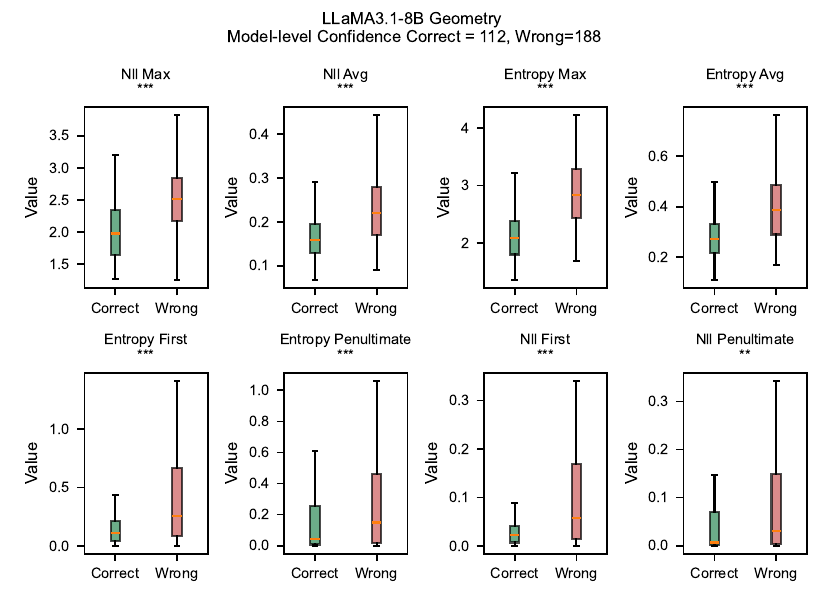}
\vspace{0.1cm} 
\includegraphics[width=0.95\linewidth]{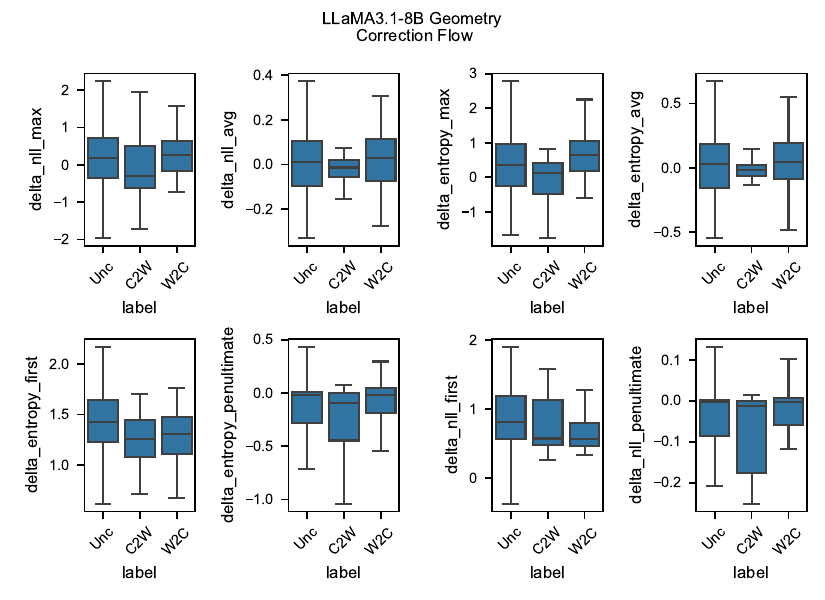}
\caption{Top: Model-level Confidence result on the Math Geometry dataset with LLaMA3.1-8B. Bottom: Correction flow with 8 deltas of confidence metrics. Unc: the answer remains unchanged between the first round and the last round, C2W: correct to wrong, W2C: wrong to correct.}
\label{fig:c-llama-geo}
\end{figure}

Figure~\ref{fig:c-llama-geo} Model-level Confidence result on the Math Geometry dataset with LLaMA3.1-8B.

\begin{figure}[!t]
\centering
\includegraphics[width=0.95\linewidth]{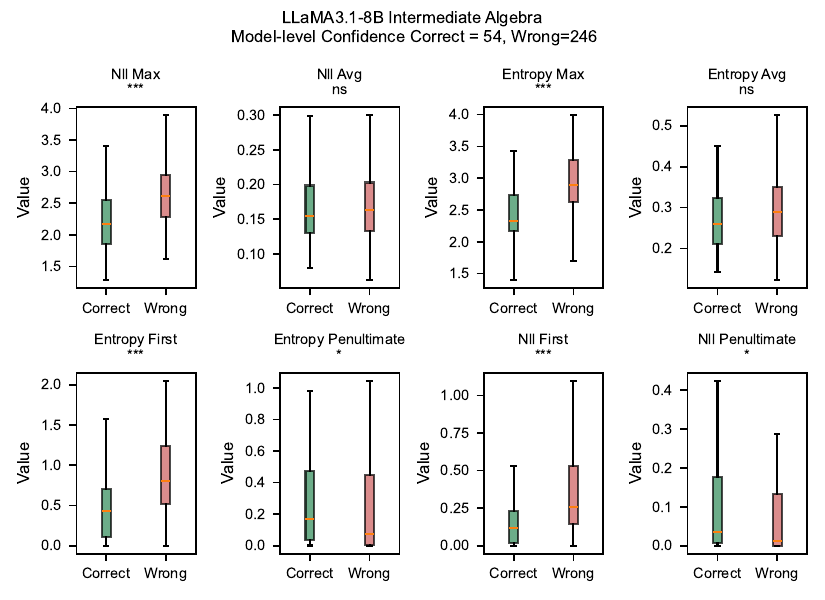}
\vspace{0.1cm} 
\includegraphics[width=0.95\linewidth]{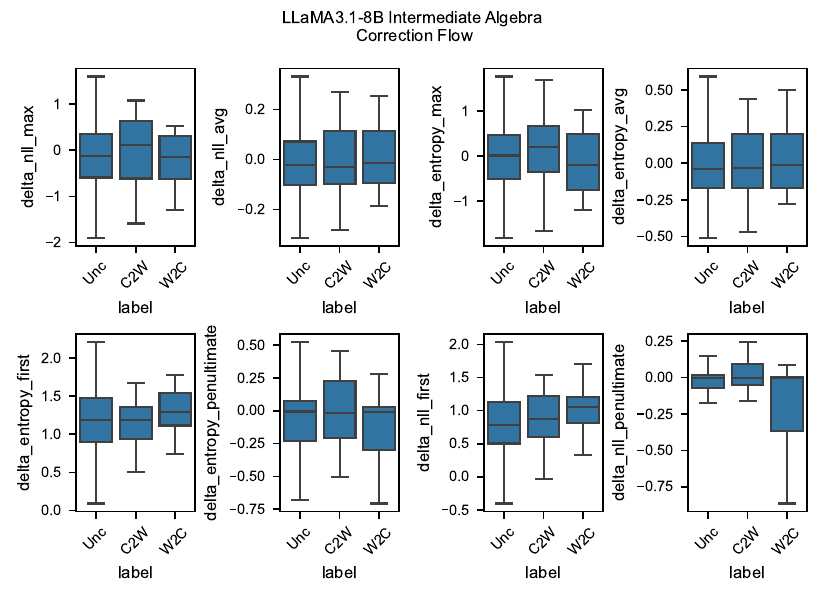}
\caption{Top: Model-level Confidence result on the Math Intermediate Algebra dataset with LLaMA3.1-8B. Bottom: Correction flow with 8 deltas of confidence metrics. Unc: the answer remains unchanged between the first round and the last round, C2W: correct to wrong, W2C: wrong to correct.}
\label{fig:c-llama-intalg}
\end{figure}

Figure~\ref{fig:c-llama-intalg} Model-level Confidence result on the Math Intermediate Algebra dataset with LLaMA3.1-8B.

\begin{figure}[!t]
\centering
\includegraphics[width=0.95\linewidth]{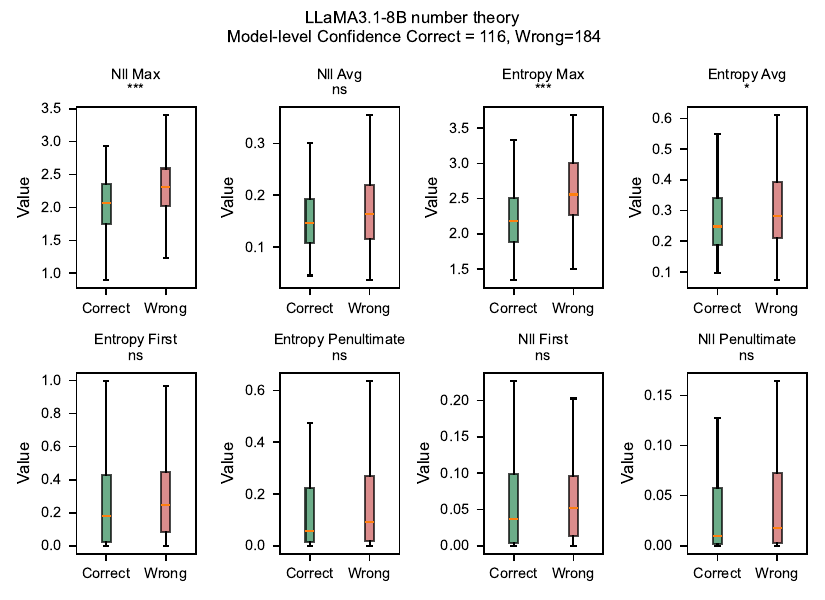}
\vspace{0.1cm} 
\includegraphics[width=0.95\linewidth]{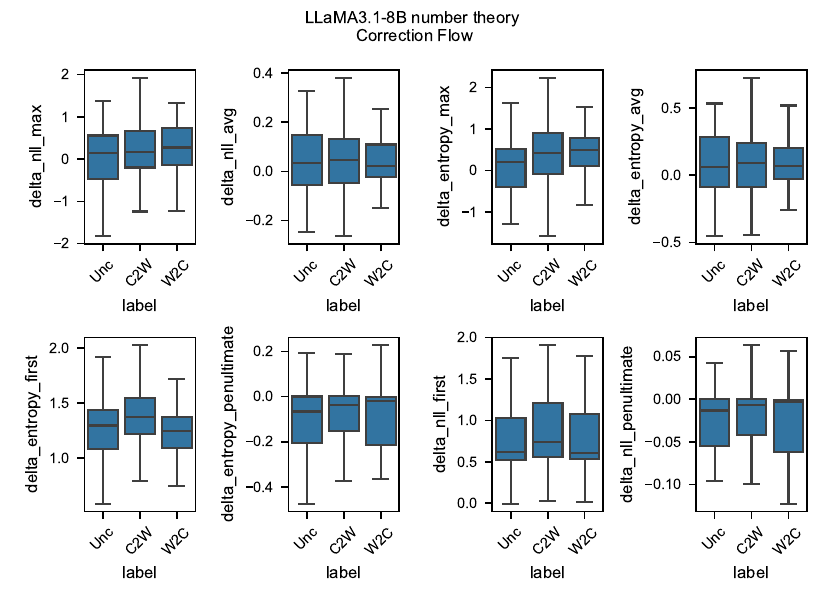}
\caption{Top: Model-level Confidence result on the Math Number Theory dataset with LLaMA3.1-8B. Bottom: Correction flow with 8 deltas of confidence metrics. Unc: the answer remains unchanged between the first round and the last round, C2W: correct to wrong, W2C: wrong to correct.}
\label{fig:c-llama-num}
\end{figure}

Figure~\ref{fig:c-llama-num} Model-level Confidence result on the Math Number Theory dataset with LLaMA3.1-8B.

\begin{figure}[!t]
\centering
\includegraphics[width=0.95\linewidth]{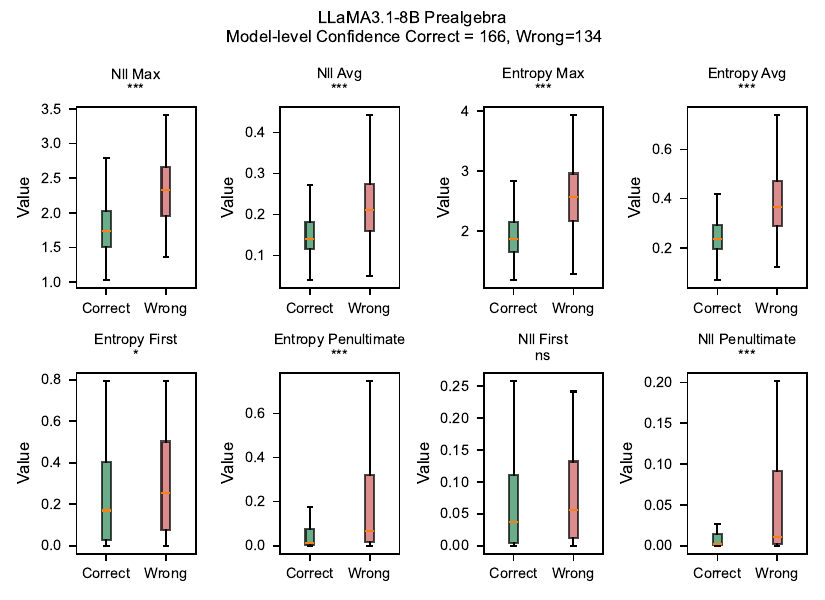}
\vspace{0.1cm} 
\includegraphics[width=0.95\linewidth]{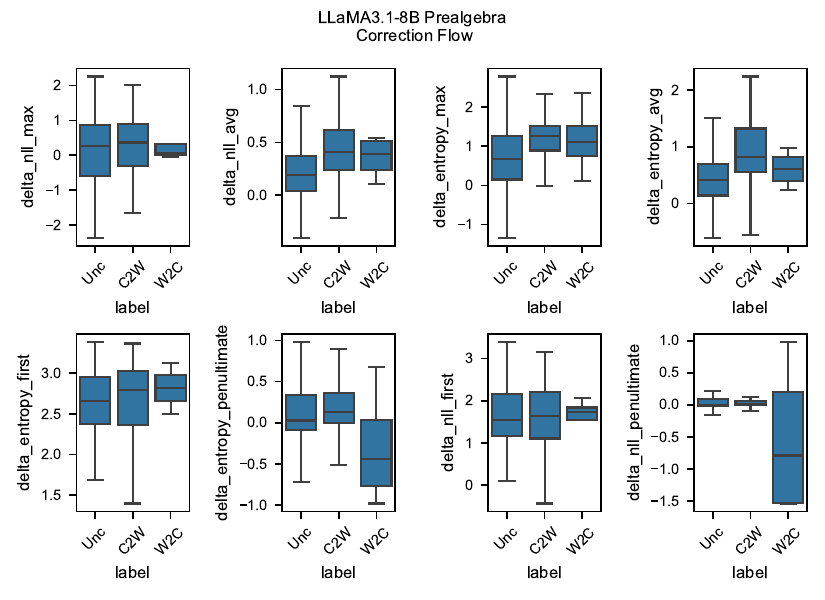}
\caption{Top: Model-level Confidence result on the Math Prealgebra dataset with LLaMA3.1-8B. Bottom: Correction flow with 8 deltas of confidence metrics. Unc: the answer remains unchanged between the first round and the last round, C2W: correct to wrong, W2C: wrong to correct.}
\label{fig:c-llama-preal}
\end{figure}

Figure~\ref{fig:c-llama-preal} Model-level Confidence result on the Math Prealgebra dataset with LLaMA3.1-8B.

\begin{figure}[!t]
\centering
\includegraphics[width=0.95\linewidth]{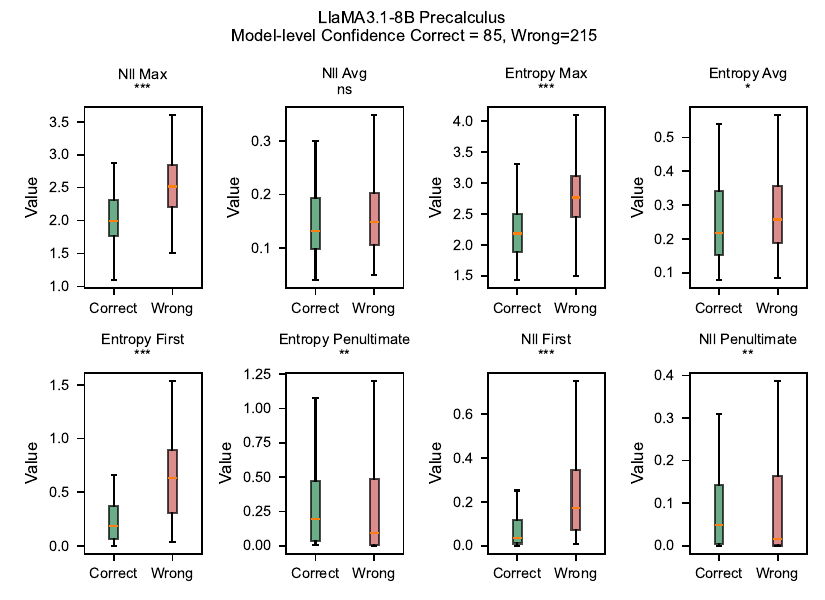}
\vspace{0.1cm} 
\includegraphics[width=0.95\linewidth]{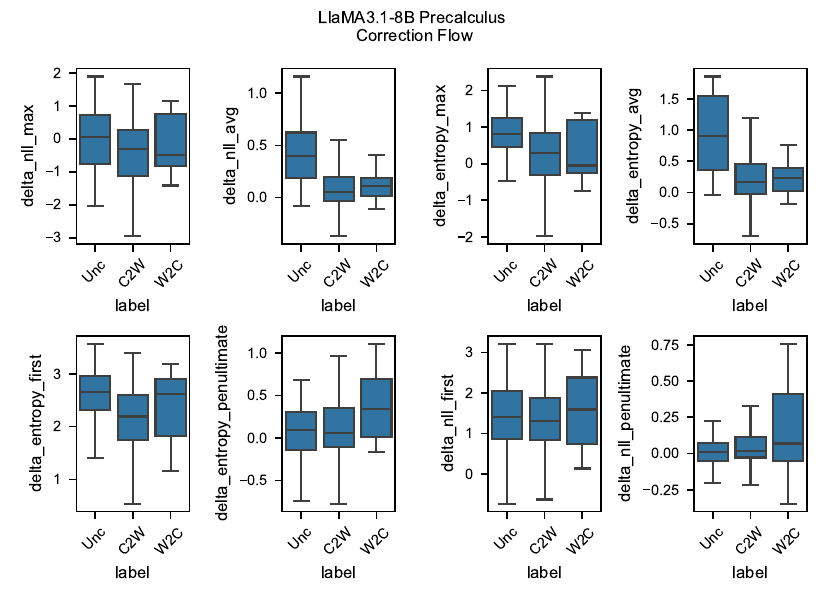}
\caption{Top: Model-level Confidence result on the Math Precalculus dataset with LLaMA3.1-8B. Bottom: Correction flow with 8 deltas of confidence metrics. Unc: the answer remains unchanged between the first round and the last round, C2W: correct to wrong, W2C: wrong to correct.}
\label{fig:c-llama-precal}
\end{figure}

Figure~\ref{fig:c-llama-precal} Model-level Confidence result on the Math Precalculus dataset with LLaMA3.1-8B.

\begin{figure}[h]
\centering
\includegraphics[width=0.95\linewidth]{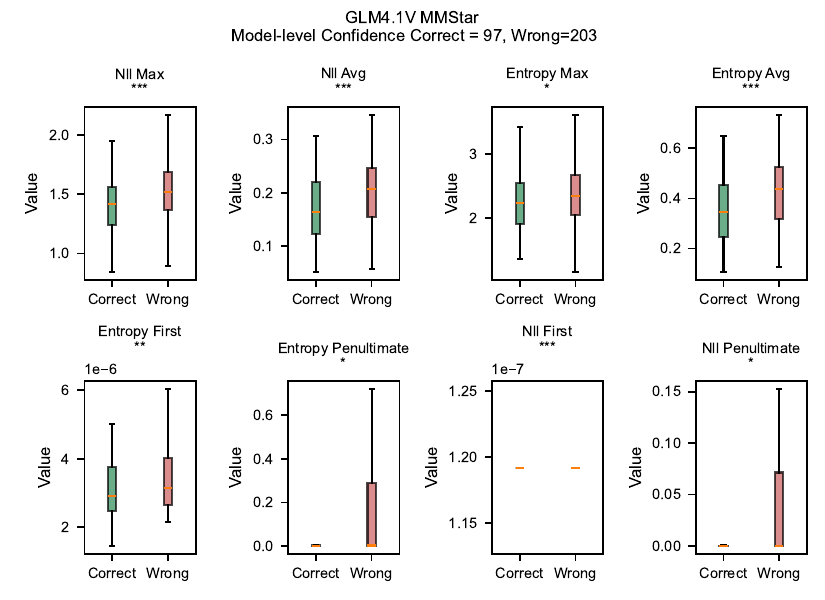}
\vspace{0.1cm} 
\includegraphics[width=0.95\linewidth]{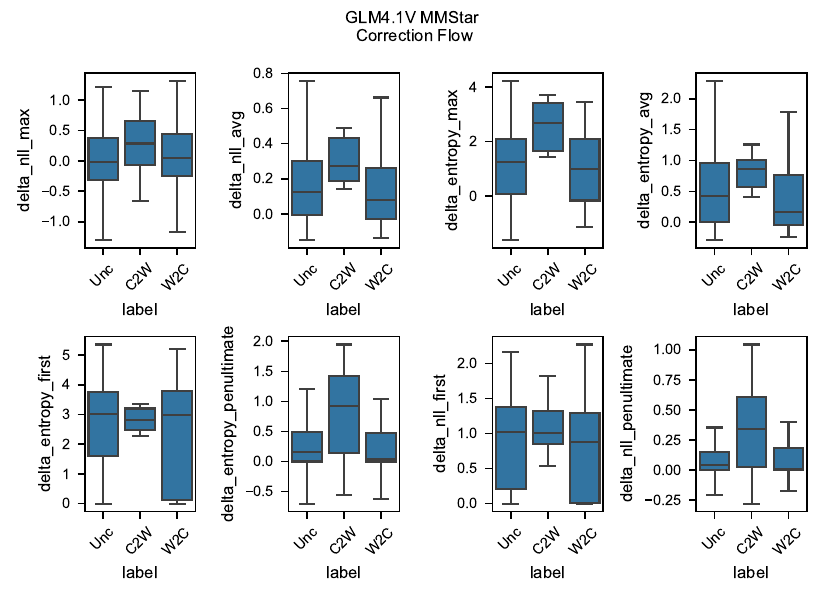}
\caption{Top: Model-level Confidence result on the MMStar dataset with GLM4.1V. Bottom: Correction flow with 8 deltas of confidence metrics. Unc: the answer remains unchanged between the first round and the last round, C2W: correct to wrong, W2C: wrong to correct.}
\label{fig:c-glm-mmstar}
\end{figure}

Figure~\ref{fig:c-glm-mmstar} Model-level Confidence result on the MMStar dataset with GLM4.1V. 

\begin{figure}[h]
\centering
\includegraphics[width=0.95\linewidth]{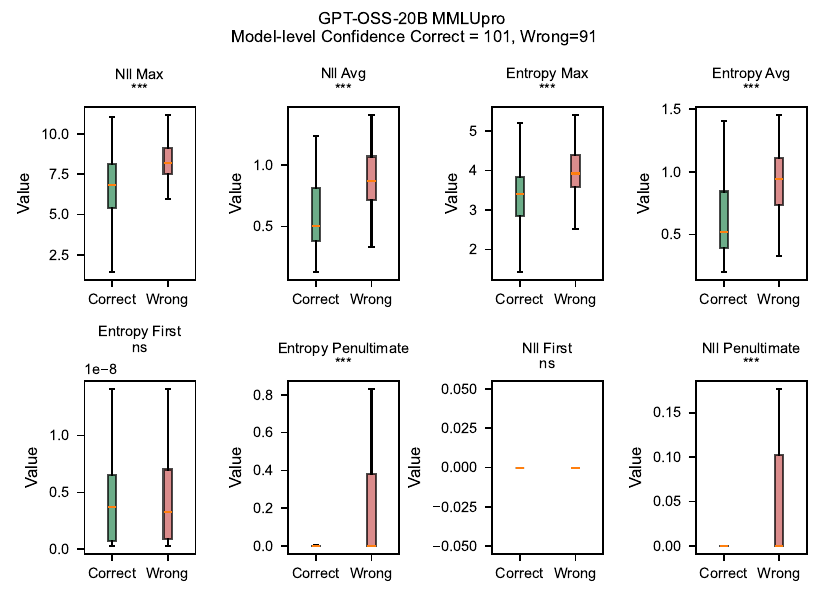}
\vspace{0.1cm} 
\includegraphics[width=0.95\linewidth]{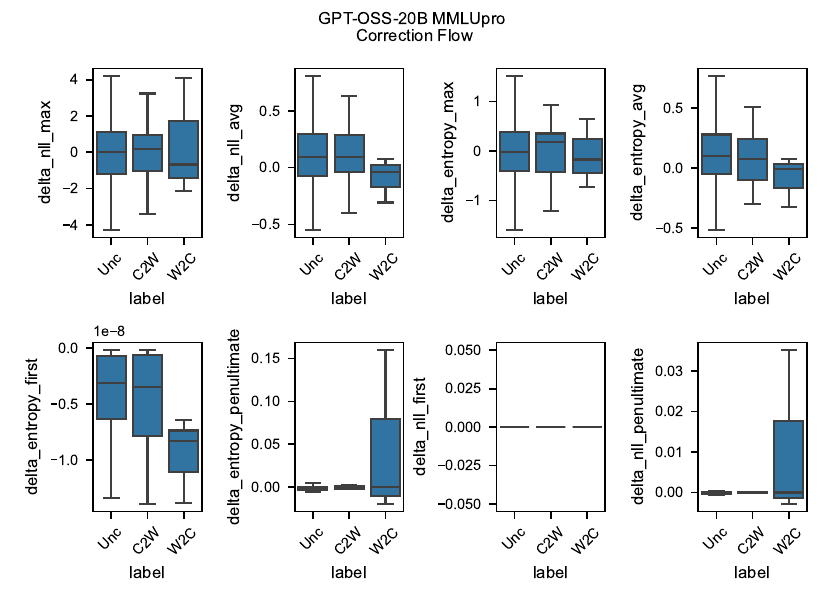}
\caption{Top: Model-level Confidence result on the MMLUpro dataset with GPT-OSS-20B. Bottom: Correction flow with 8 deltas of confidence metrics. Unc: the answer remains unchanged between the first round and the last round, C2W: correct to wrong, W2C: wrong to correct.}
\label{fig:c-gpt-mmlupro}
\end{figure}

Figure~\ref{fig:c-gpt-mmlupro}  Model-level Confidence result on the MMLUpro dataset with GPT-OSS-20B. Bottom: Correction flow with 8 deltas of confidence metrics.

In addition, a number of model-level early exit cases are provided here to show the confident and overconfident cases. It can be observed that the model partially analyzes the problem in overconfident cases.

Figure~\ref{fig:a-ml1}  Examples of model-level early exit cases in the MMLUpro dataset.

\begin{figure}[h]
\centering
\includegraphics[page=3, width=0.95\linewidth]{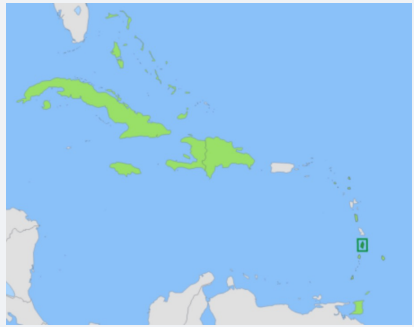}
\caption{Examples of model-level early exit cases on ScienceQA datasets. The exit questions are correctly answered in principle.}
\label{fig:a-ml2}
\end{figure}

Figure~\ref{fig:a-ml2} Examples of model-level early exit cases in the ScienceQA dataset.

Figure~\ref{fig:a-ml3} Examples of model-level early exit cases in the Math dataset.

Figure~\ref{fig:a-ml4}Examples of model-level early exit cases in the GPQA dataset.

\begin{figure}[!t]
\centering
\includegraphics[page=1, width=0.95\linewidth]{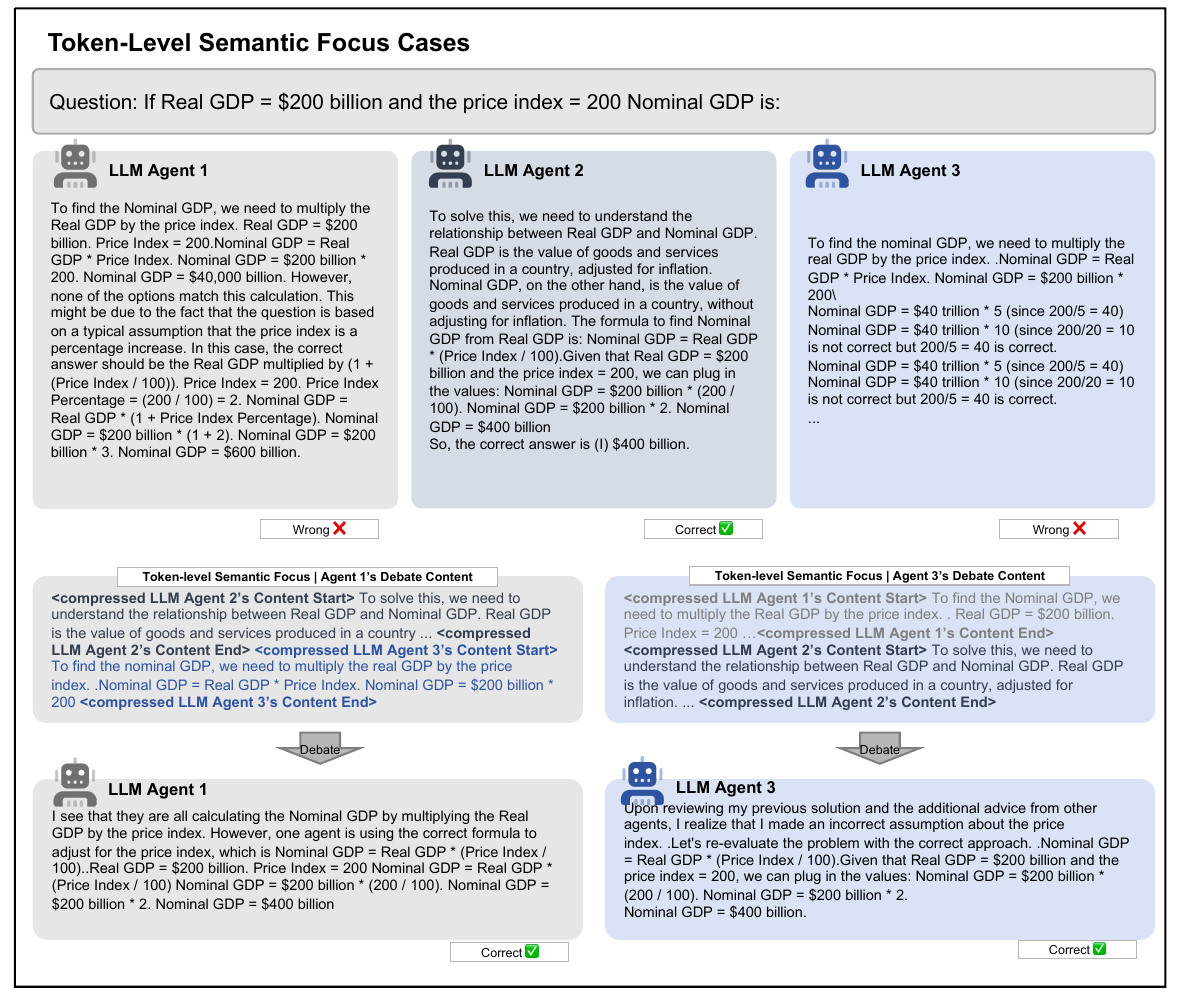}
\caption{Visualization of token-level semantic focus helped agents correct their answers during the SID. Specifically, SID invokes a compressed debate round, highlighting disagreement-relevant spans across debate contents.  Agent 1 and Agent 3 iteratively revise their reasoning based on focused inputs, ultimately correcting earlier errors and converging on the correct answer.}
\label{fig:a-tl2}
\end{figure}

Subsequently, Figure~\ref{fig:a-tl1} and Figure~\ref{fig:a-tl2} display that the token-level semantic focus module compresses the contents and assists agents in correcting their answers.

% \clearpage

\begin{figure}[!t]
\centering
\includegraphics[width=0.98\linewidth]{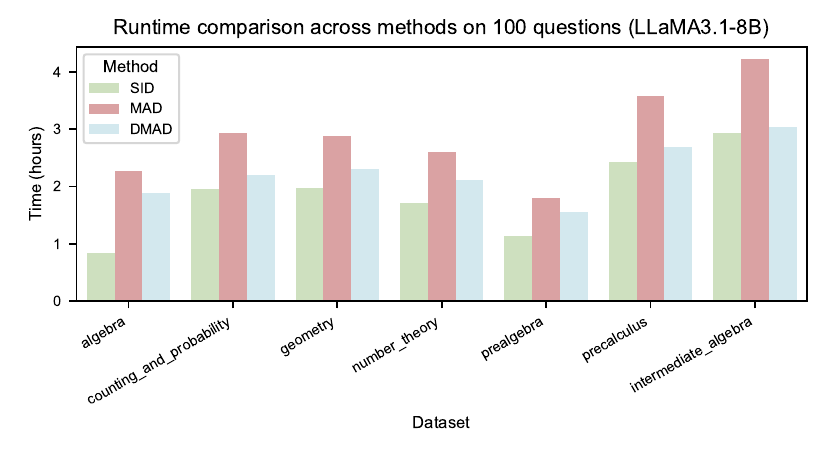}
\caption{Running time comparison on math datasets with LLaMA3.1-8B, running on single A100 80GB GPU}
\label{fig:runtime}
\end{figure}

\begin{figure}[h]
\centering
\includegraphics[width=0.95\linewidth]{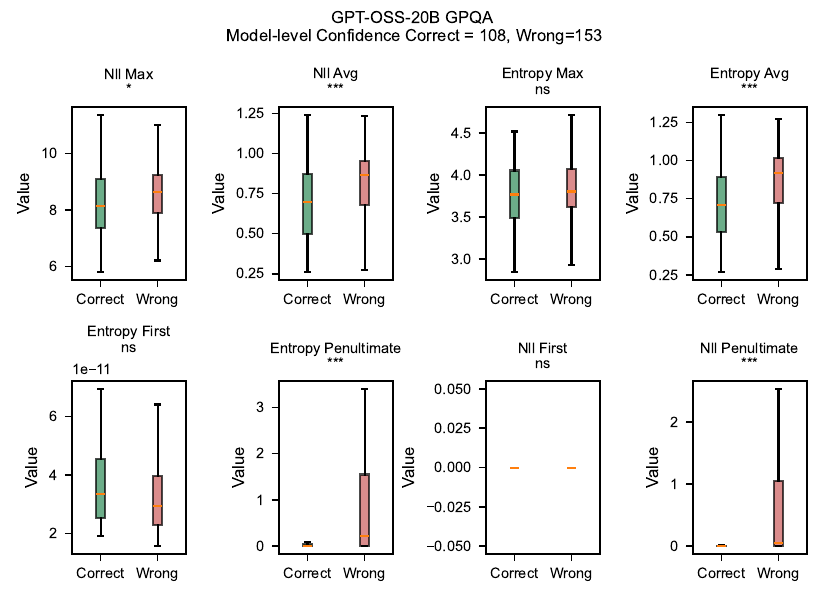}
\vspace{0.1cm} 
\includegraphics[width=0.95\linewidth]{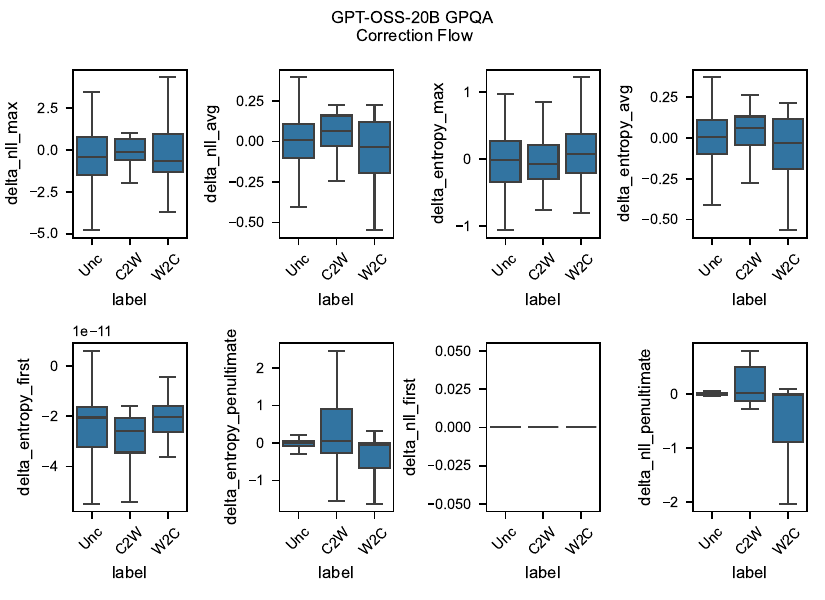}
\caption{Top: Model-level Confidence result on the GPQA dataset with GPT-OSS-20B. Bottom: Correction flow with 8 deltas of confidence metrics. Unc: the answer remains unchanged between the first round and the last round, C2W: correct to wrong, W2C: wrong to correct.}
\label{fig:c-gpt-gpqa}
\end{figure}

\begin{figure}[h]
\centering
\includegraphics[page=2, width=0.95\linewidth]{fig/a4_appdenix-case-v4.pdf}
\caption{Examples of model-level early exit cases on MMLUpro datasets. The exit questions are correctly answered in principle.}
\label{fig:a-ml1}
\end{figure}

\begin{figure}[h]
\centering
\includegraphics[page=4, width=0.95\linewidth]{fig/a4_appdenix-case-v4.pdf}
\caption{Examples of model-level early exit cases on Math datasets. The bottom overconfident case shows the model partially analyzes the question, potentially leading to a high confidence.}
\label{fig:a-ml3}
\end{figure}

\begin{figure}[h]
\centering
\includegraphics[page=5, width=0.95\linewidth]{fig/a4_appdenix-case-v4.pdf}
\caption{Examples of model-level early exit cases on GPQA datasets, where we did not find negative cases (\emph{i.e.,} the model is overconfident).}
\label{fig:a-ml4}
\end{figure}

\begin{figure}[h]
\centering
\includegraphics[page=6, width=0.95\linewidth]{fig/a4_appdenix-case-v4.pdf}
\caption{Detailed example of token-level semantic focus. There is content from two agents in the figure. The first agent's answer was largely adopted, whereas the second agent's answer was largely cropped. <comments> shows places that were adaptively compressed by the model's attention mechanism.}
\label{fig:a-tl1}
\end{figure}

\end{document}